\documentclass{article}
\usepackage{PRIMEarxiv}
\usepackage[utf8]{inputenc} 
\usepackage[T1]{fontenc}    
\usepackage{hyperref}       
\usepackage{url}            
\usepackage{booktabs}       
\usepackage{amsfonts}       
\usepackage{nicefrac}       
\usepackage{microtype}      
\usepackage{lipsum}
\usepackage{makecell}
\usepackage{fancyhdr}       
\usepackage{graphicx}       
\usepackage{amsmath}
\usepackage{array}
\usepackage{booktabs}
\usepackage{bbm}
\usepackage{float}

\title{Self-Normalizing Foundation Model for Enhanced Multi-Omics Data Analysis in Oncology
}

\author{
    Asim Waqas$*^{,\dagger}$, Aakash Tripathi $^{\dagger}$, Sabeen Ahmed $^{\dagger}$ \\
  Departments of Cancer Epidemiology and Machine Learning/ EE Department  \\
  Moffitt Cancer Center \& Research Institute/ University of South Florida \\
  \texttt{\{Asim.Waqas, Aakash.Tripathi, Sabeen.Ahmed\}@moffitt.org} \\
   \And
  Ashwin Mukund $^{\dagger}$ \\
  Department and Machine Learning \\
  Moffitt Cancer Center \& Research Institute \\
  \And
  Joseph O. Johnson \\
  Analytic Microscopy Core Facility \\
  Moffitt Cancer Center \& Research Institute \\
  \And
  Hamza Farooq \\ 
  Center for Magnetic Resonance Research \\
  University of Minnesota \\
  \And
  Matthew B. Schabath \\
  Departments of Cancer Epidemiology and Thoracic Oncology \\
  Moffitt Cancer Center \& Research Institute \\
  \And
  Paul Stewart \\
  Department of Biostatistics and Bioinformatics \\
  Moffitt Cancer Center \& Research Institute \\
  \And
  Mia Naeini \\
  Department of Electrical Engineering \\
  University of South Florida \\
  \And
  Ghulam Rasool \\
  Department of Machine Learning/ EE Department  \\
  Moffitt Cancer Center \& Research Institute/ University of South Florida \\
  \texttt{Ghulam.Rasool@moffitt.org} \\
}

\begin{document}
\maketitle

\begin{abstract}
Multi-omics research has enhanced our understanding of cancer heterogeneity and progression. Investigating molecular data through multi-omics approaches is crucial for unraveling the complex biological mechanisms underlying cancer, thereby enabling more effective diagnosis, treatment, and prevention strategies. However, predicting patient outcomes through the integration of all available multi-omics data is still an under-study research direction. Here, we present SeNMo (Self-normalizing Network for Multi-omics), a foundation model that has been trained on multi-omics data across 33 cancer types. SeNMo is particularly efficient in handling multi-omics data characterized by high-width (many features) and low-length (fewer samples) attributes. We trained SeNMo for the task of overall survival of patients using pan-cancer multi-omics data involving 33 cancer sites from the Genomics Data Commons (GDC). The training multi-omics data includes gene expression, DNA methylation, miRNA expression, DNA mutations, protein expression modalities, and clinical data. SeNMo was validated on two independent cohorts: Moffitt Cancer Center and CPTAC lung squamous cell carcinoma. We evaluated the model’s performance in predicting patient’s overall survival using the concordance index (C-Index). SeNMo performed consistently well in the training regime, reflected by the validation C-Index of $0.76$ on GDC's public data. In the testing regime, SeNMo performed with a C-Index of $0.758$ on a held-out test set. The model showed an average accuracy of $99.8\%$ on the task of classifying the primary cancer type on the pan-cancer test cohort. SeNMo demonstrated robust performance on the classification task of predicting the primary cancer type of patients. SeNMo further demonstrated significant performance in predicting tertiary lymph structures from multi-omics data, showing generalizability across cancer types, molecular data types, and clinical endpoints. We believe SeNMo and similar models are poised to transform the oncology landscape, offering hope for more effective, efficient, and patient-centric cancer care.
\end{abstract}


\section{Introduction}

Across the cancer care continuum, from screening, diagnosis, treatment, to survivorship, vast amounts of standard-of-care data are collected from patients. In cancer research, the volume and diversity of data further expand, providing distinct and complementary views of the disease \cite{jiang2022big}. For instance, radiological images capture structural and functional information at the organ and sub-organ levels, histopathology slides offer morphological, cellular, and tissue-level insights, clinical and Electronic Health Records (EHR) encapsulate patient history, treatment plans, and outcomes, while molecular data—such as genomics, transcriptomics, proteomics, and metabolomics—reveal the underlying biological mechanisms driving cancer progression and treatment response \cite{bera2022predicting,krithiga2021breast,morin2021artificial,chatsirisupachai2021integrative}. Studying cancer from a multimodal perspective is essential for comprehensive understanding and for developing effective, personalized treatment strategies \cite{hanahan2011hallmarks, acosta2022multimodal}.

\textbf{Multimodal and multi-omics data}. The advancement of technologies to record, process, and store molecular data has significantly propelled cancer research \cite{chatsirisupachai2021integrative}. High-throughput sequencing technologies, along with sophisticated bioinformatics tools and computational algorithms, have ushered in an era of ``omics" \cite{NGS}. Multi-omics, a subset of multimodal data, specifically refers to the integrated analysis of various molecular modalities, including genomics, transcriptomics, proteomics, and metabolomics \cite{waqas2023multimodal}. Multi-omics provides a comprehensive understanding of the biological processes and molecular mechanisms underlying cancer \cite{zhao2024tutorial}. By combining different layers of molecular data, multi-omics transcends the limitations of single-omic studies, which often provide only a partial view of the disease. It illustrates how various molecular components, such as DNA mutations, protein expression, and RNA expression, interact within the complex biological network of cancer \cite{hasin2017multi}.


\textbf{Pan-cancer perspective}. Cancer research can be approached from two primary perspectives: individual cancer studies and pan-cancer studies. Individual cancer studies focus on a specific type of cancer, delving deep into its unique molecular and genetic characteristics, allowing for the development of highly targeted therapies and personalized treatment plans. Studying individual cancers has shown significant benefits in understanding specific pathways and therapeutic responses. Conversely, pan-cancer studies analyze commonalities and differences across multiple cancer types, uncovering shared molecular mechanisms and genetic alterations. This approach reveals broader patterns and potentially identifies universal biomarkers or therapeutic targets applicable across different cancers, enhancing our holistic understanding of the disease \cite{underwood2020pan}. The pan-cancer perspective has uncovered universal cancer vulnerabilities, detailed pathway alterations for cross-cancer diagnostics and treatments, and revealed shared oncogenic pathways and mutation patterns, leading to new clinically useful insights \cite{hu2020multi, sanchez2018oncogenic, hoadley2018cell, underwood2020pan}. Furthermore, pan-cancer studies have identified key molecular signatures that can predict response to immunotherapy across diverse tumor types, demonstrating the wide-reaching clinical significance of the pan-cancer approach \cite{thorsson2018immune, li2023pan}. In this article, we focus on the pan-cancer perspective, emphasizing its potential to generate overarching insights that could lead to more comprehensive and versatile cancer treatment strategies.


\textbf{Existing landscape of pan-cancer multi-omics analysis}.
Traditionally, multimodal, multi-omics, and pan-cancer studies have been conducted through a variety of techniques and methods that leverage advanced computational, bioinformatics, statistical, machine learning, and deep learning approaches to integrate and interpret complex oncology datasets. Data integration techniques in multi-omics are generally categorized into supervised, weakly supervised, and unsupervised methods. These methods can be further sub-categorized into (1) feature extraction (selection, extraction, and dimensionality reduction), (2) feature engineering (transformation, dimensionality reduction, data normalization, simplification, noise reduction, and alignment), (3) network-based methods (e.g., patient similarity networks, patient-drug networks, drug-drug networks), (4) clustering (e.g., grouping similar samples, stratification, feature selection, biological module grouping), (5) factorization (e.g., feature decomposition, multiple kernel learning, Bayesian consensus, similarity network fusion, non-negative matrix factorization), and (6) deep learning techniques (e.g., Convolutional Neural Networks (CNNs), Multilayer Perceptions (MLPs), Recurrent Neural Networks (RNNs), Transformers, Graph Neural Networks (GNNs)) \cite{acharya2024comprehensive, waqas2023multimodal, ahmed2023transformers, waqas2021brain}. Deep learning, a subset of machine learning characterized by neural networks with many layers, has transformed the study of high-dimensional, low-sample molecular data \cite{ahmed2022failure, waqas2022exploring}. With its capacity to model complex, non-linear relationships and handle vast datasets, deep learning has proven adept at uncovering patterns that traditional statistical and machine learning models may not identify. Numerous reviews in existing literature provide in-depth analysis of various pan-cancer, multimodal, and multi-omics research efforts \cite{lipkova2022artificial, boehm2022harnessing, he2023artificial, steyaert2023multimodal, waqas2023multimodal, waqas2024bio24, tripathi2024multimodal, MINDS}.

A significant advancement in the field is the use of self-normalizing neural networks for pan-cancer classification. A study leveraging copy number variation data from The Cancer Genome Atlas (TCGA) for lung adenocarcinoma (LUAD), ovarian cancer (OV), liver hepatocellular carcinoma (LIHC), and breast cancer (BRCA) demonstrated that feature selection is crucial for managing high-dimensional data in disease categorization \cite{li2020pan}. The self-normalizing model for pan-cancer classification yielded superior accuracy and macro F1 scores compared to a traditional random forest algorithm \cite{li2020pan}. Complementing this approach, an integrative analysis that combined histology-genomic data using multimodal deep learning provided broad-spectrum insights into cancer biology \cite{chen2022pan}. Using an extensive dataset from TCGA encompassing 14 cancer types, a deep learning multimodal fusion model outperformed an attention-based multiple-instance learning model and a self-normalizing network, demonstrating the benefits of integrative analytics over single data type analyses \cite{chen2022pan}. Emphasizing multi-omics data integration, DeepProg—an ensemble framework that combines deep learning and machine learning—achieved high performance in prognosis prediction \cite{poirion2021deepprog}. By processing RNA-Seq, miRNA sequencing, and DNA methylation data for 32 cancer types from TCGA, DeepProg excelled in predicting survival subtypes and risk stratification \cite{poirion2021deepprog}.

Khadirnaikar \textit{et al}. identified novel subgroups with similar molecular characteristics by combining different machine learning and deep learning models \cite{khadirnaikar2023integration}. By reducing the dimensionality of multi-omics features (e.g., mRNA, miRNA, DNA methylation, protein expression) and applying multiple classifiers, this approach successfully identified subgroups across 33 tumor types. The authors argued that the number of samples should be proportional to the number of features for optimal predictive power of a learning model \cite{khadirnaikar2023integration}. Another study used four types of -omics data (gene expression, miRNA expression, protein expression, and DNA methylation) for two datasets (TCGA-BLCA, TCGA-LGG) to predict progression-free interval and overall survival (OS) through a multiview factorization autoencoder \cite{ma2019integrate}. The identification of pan-cancer prognostic biomarkers using integrated multi-omics data (including DNA methylation, gene expression, somatic copy number alteration, and miRNA expression) across 13 cancers highlighted the power of statistical and bioinformatics methods for discovering survival-related genes \cite{zhao2020identification}. The predictive capability of multi-omics data was also evident in non-small cell lung cancer survival prediction, where combining five modalities—miRNA, mRNA, DNA methylation, long non‑coding RNA, and clinical data—resulted in a superior concordance index (C-Index) compared to individual modalities \cite{ellen2023autoencoder}.

The advantage of multimodal data fusion for predicting OS was quantified across various cancer stages and types, with fused models exhibiting higher average C-Index compared to machine learning and bioinformatics methods \cite{nikolaou2024quantifying}. This approach combined clinical features with genomic, transcriptomic, and proteomic data in oncological prognostics across 33 cancer types \cite{nikolaou2024quantifying}. A deep learning-based clustering method called MCluster-VAEs achieved superior performance in subtype discovery using multi-omics data (e.g., mRNA, miRNA, DNA methylation, CNA) across 32 cancer types \cite{rong2022mcluster}. The decoupled contrastive learning model DEDUCE employed a multi-head attention decoupled contrastive learning approach for subtype clustering through multi-omics data consisting of gene expression, DNA methylation, and miRNA expression across five cancer types (BRCA, GBM, SARC, LUAD, STAD) \cite{pan2023multi}. The authors of DEDUCE utilized a multi-head attention encoder network for cancer subtype discovery \cite{pan2023multi}.

\textbf{Limitations of the State-of-the-art Methods}.
Although valuable for their intended tasks, the above-mentioned methods often struggle to fully capture the complexity and heterogeneity of cancer due to inherent limitations in handling and interpreting vast, multidimensional datasets. Dimensionality reduction methods such as principal component analysis or t-distributed stochastic neighbor embedding can inadvertently discard subtle yet crucial biological nuances that are pivotal for understanding disease mechanisms \cite{jia2022feature}. Learning-based dimensionality reduction methods, such as those utilizing deep learning, face challenges including limited discriminative and interpretive capabilities of extracted features, lack of consensus on the balance between the number of network layers and the number of neurons per layer, and limitations in handling or recovering missing data \cite{jia2022feature}.

Similarly, feature selection and learning-based feature engineering, despite being effective in identifying key predictors, can introduce biases and create models that are overly tailored to specific features within training datasets \cite{krawczuk2016feature, yang2023causal}. This bias undermines generalizability across diverse datasets or real-world clinical settings \cite{krawczuk2016feature, yang2023causal}. Additionally, these methods frequently face challenges in ensuring consistent performance across varied patient populations and biological conditions, limiting their broader clinical utility. Thus, while these techniques are instrumental in advancing cancer research, they underscore the need for more robust and generalizable frameworks capable of accurately predicting endpoints across diverse cancer types and data modalities.

Recently, a new class of deep learning models called foundation models, which include large language models and vision-language models, have been introduced by training on large multimodal datasets \cite{waqas2023revolutionizing, hartsock2024vision}. These models have demonstrated a strong ability to generalize across different tasks when provided with diverse and substantial training data \cite{waqas2023revolutionizing}. By leveraging extensive and varied datasets, these models are able to capture a broad spectrum of patterns and nuances, allowing for flexible and effective application across different contexts. Key conclusions from the successes of foundation models relevant to this study are:
\begin{enumerate}
        \item \textbf{Extensive training data}: 
        Foundation models are trained on massive datasets encompassing diverse domains and modalities. This extensive training helps models develop a robust understanding of complex patterns and relationships within data. For example, Generative Pre-trained Transformers (GPT) \cite{brown2020language} and Bidirectional Encoder Representations from Transformers (BERT) \cite{devlin2018bert} have been shown to excel across various natural language processing tasks, from translation to sentiment analysis, due to exposure to large and varied textual datasets during training \cite{hartsock2024vision, waqas2023revolutionizing}.
        
        \item \textbf{Cross-modal learning}: 
        Vision-language models integrate visual and textual information, enabling comprehensive understanding. Models like Contrastive Language-Image Pre-Training (CLIP) \cite{radford2021learning} and Vision-and-Language BERT (ViLBERT) \cite{lu2019vilbert} correlate images and text, allowing them to perform tasks such as image captioning or visual question answering with high accuracy. This cross-modal processing and synthesizing capability enhances adaptability to new tasks beyond their original training scope \cite{hartsock2024vision}.
               
        \item \textbf{Generalization across tasks}: 
        Foundation models exhibit impressive generalization across diverse tasks with minimal task-specific tuning \cite{waqas2023revolutionizing}. Once trained, they can switch between tasks like text classification, summarization, and complex reasoning without extensive retraining. This adaptability is largely due to their training datasets' comprehensive and diverse nature, which provides a rich background against which the models can evaluate new problems \cite{waqas2023revolutionizing, waqas2023multimodal, hartsock2024vision}.
    \end{enumerate}

\begin{figure}[ht!]%
    \centering
    \includegraphics[width=1.0\textwidth]{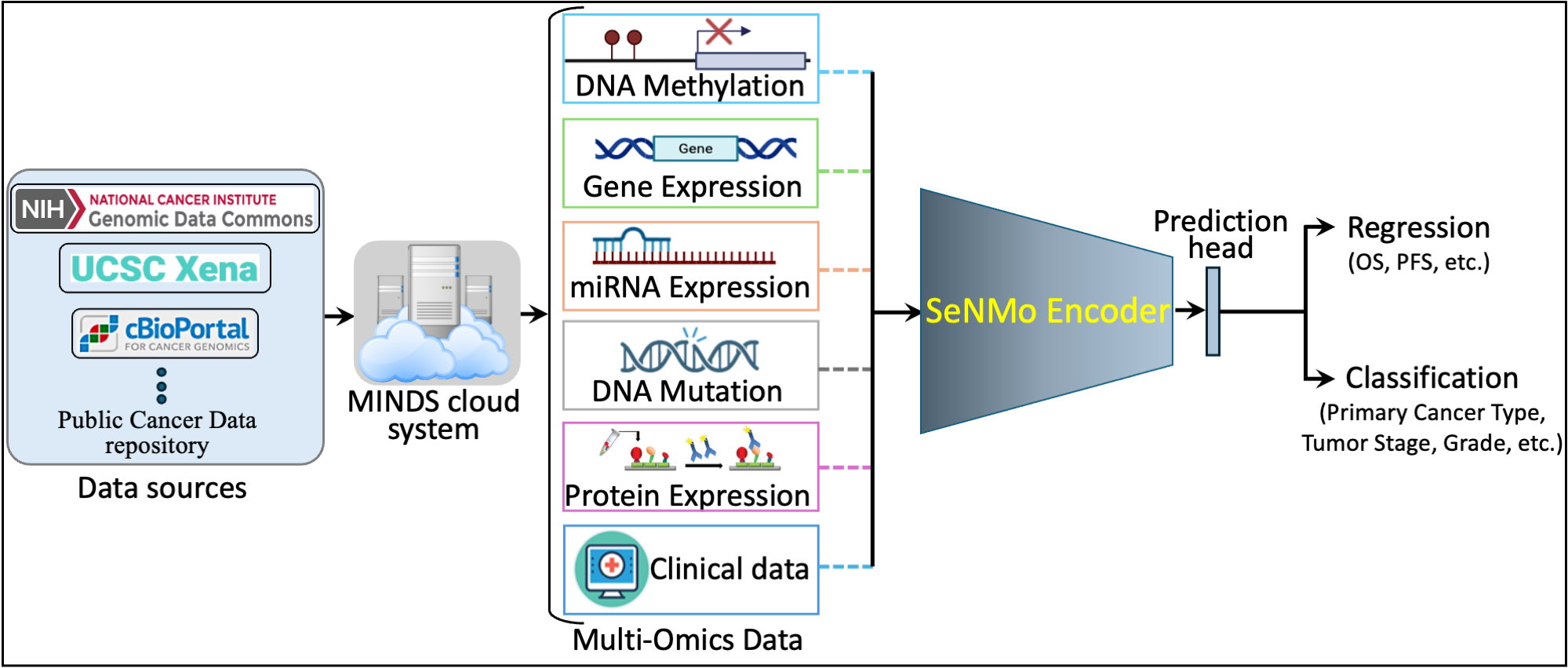} %
    \caption{Overview of the SeNMo model. The data from public sources are collected using MINDS \cite{MINDS} and curated to develop the multimodal dataset for SeNMo training. MINDS is a metadata framework for fusing publicly available data sources like TCGA-GDC and UCSC Xena Portal into machine learning-ready format \cite{TCGA, UCSCXena, MINDS}. The dataset is preprocessed and fed to the self-normalizing deep learning encoder network that learns underlying sub-visual patterns from cross-modality, pan-cancer data. The learned encoder weights are later used for different downstream tasks (with or without fine-tuning), such as predicting the overall survival (OS), progression-free survival (PFS), cancer subtype classification, grading, or tertiary lymph structures (TLS) ratio.}
    \label{fig:SeNMo}
\end{figure}

The establishment of large-scale biological databases and data repositories, such as the National Cancer Institute's TCGA \cite{TCGA} and the Clinical Proteomic Tumor Analysis Consortium (CPTAC) \cite{CPTAC}, hold vast amounts of multi-omics cancer data that are readily available for disease analysis. Despite numerous efforts, existing literature lacks a foundation model trained on multi-omics pan-cancer data. scGPT is a foundation model trained for single-cell sequencing data comprising 33 million cells \cite{cui2024scgpt}. The SAMMS model was trained on two cancer types (TCGA’s LGG and KIRC) using patient-level data (age, gender), gene expression, CNV, miRNA, and WSI \cite{zhu2023samms}. The RNA Foundation Model (RNA-FM) was trained on 23 million non-coding RNA sequences \cite{chen2022interpretable}. PATH-GPTOMIC utilized CNV, genomic mutations, bulk RNA Seq, and WSI data to predict survival outcomes for two datasets (TCGA-GBMLGG, TCGA-KIRC) \cite{wang2024path}. The absence of a pan-cancer, multi-omics foundation model can be attributed to challenges such as data complexity, heterogeneity, limited comprehensive datasets, specificity of analytical methods, and large computational demands. To address these challenges, we propose a multi-omics, pan-cancer framework with minimal preprocessing, introducing a foundation model called the 'Self-Normalizing Deep Learning Model for Multi-Omics' (SeNMo). SeNMo has been trained on six data modalities, including clinical, gene expression, miRNA expression, DNA methylation, DNA mutations, and reverse-phase protein array (RPPA) expression data across 33 cancer types. We have evaluated SeNMo for generalization, scalability, emergence, expressivity, and compositionality, which are essential traits for a true foundation model \cite{alfasly2023foundation, waqas2023revolutionizing}. We evaluated SeNMo's generalization capability to unseen datasets and across different tasks such as OS prediction, primary cancer classification, and tertiary lymph structures (TLS) ratio prediction. Figure \ref{fig:SeNMo} presents the overview of our framework.

This work offers the following contributions:
\begin{enumerate}
    \item We present an oncology data analysis using molecular correlates of patient prognosis across 33 cancer types, addressing both disease-wide and individual patient levels.
    \item We created a multi-omics, pan-cancer framework with minimal and essential preprocessing steps, eliminating the need for complex, custom-engineered methods, thereby allowing a greater focus on the learning aspect.
    \item We developed a foundation model capable of generalizing across different tasks and to unseen data through fine-tuning.
    \item Our findings indicate that MLP-based networks are highly susceptible to catastrophic forgetting. We demonstrate that fine-tuning should involve a fraction of the epochs ($\leq30$), while adjusting the learning rate, weight decay, and dropout to fractionally update all layers of the trained model.
    \item The SeNMo framework represents the first initiative to analyze 33 cancer types using six molecular data modalities: clinical data, gene expression, miRNA expression, DNA methylation, DNA mutations, and protein expression.
    \item We present the first effort to predict tertiary lymph structures (TLS) ratio from multi-omic data only.
    \item We provide the latent feature vectors (embeddings) learned by SeNMo as an open-access vector database system, HoneyBee, available through Hugging Face and GitHub.
\end{enumerate}

\section{Materials and Methods}
    \subsection{Datasets}
        \subsubsection{Data Acquisition}
        TCGA houses one of the largest collections of high-dimensional multi-omics datasets, comprising over 20,500 individual tumor samples from 33 different cancer types \cite{TCGA}. The available data includes high-throughput RNA sequencing (RNA-Seq), DNA sequencing (DNA-Seq), microRNA sequencing (miRNA-Seq), single nucleotide variants, copy number variations, DNA methylation, and reverse-phase protein array (RPPA) data \cite{TCGA}. Building cohorts from this diverse data, spanning multiple formats, modalities, and systems, presents significant challenges. To curate and establish patient cohorts, we utilized our previously developed Multimodal Integration of Oncology Data System (MINDS), a metadata framework designed to fuse data from publicly available sources like TCGA-GDC and UCSC Xena Portal into a machine learning-ready format \cite{TCGA, UCSCXena, MINDS}. MINDS is freely accessible to the cancer research community, and has been integrated into the SeNMo framework to enhance its usability and benefit to researchers. For training, validation, and testing, we used pan-cancer data from TCGA and Xena, covering 33 cancer types, as summarized in Table \ref{Tab:features}. We further fine-tuned the model using data from the CPTAC-LSCC \cite{CPTAC-LSCC} and Moffitt's LSCC datasets \cite{stewart2019proteogenomic} to evaluate the generalizability and transfer learning capabilities of SeNMo.
        
        \subsubsection{Data Modalities}
        From the 13 available multi-omic modalities present in each cancer dataset, we selected gene expression (RNAseq), DNA methylation, miRNA stem-loop expression, RPPA data, DNA mutation, and clinical data. These modalities were chosen based on their frequent use in cancer studies due to their direct relevance to the fundamental processes of cancer progression, as well as their diagnostic and prognostic capabilities \cite{sarhadi2022molecular, li2023pan}. They offer direct insights into key biological processes fundamental to cancer progression, making them extremely valuable for uncovering the molecular mechanisms driving the disease \cite{sarhadi2022molecular}. Furthermore, these selected modalities provide robust predictive and prognostic information, and their integration gives a holistic view of a tumor's multi-omic profile \cite{chen2021moving, sarhadi2022molecular, li2023pan}. Importantly, each modality had a consistent number of features across all cancer types, which facilitated the development of a standardized data preprocessing pipeline for pan-cancer studies. Below is a brief description of each data modality considered in this study, followed by the preprocessing steps used to select features for training the SeNMo model.

            \begin{enumerate}
                \item \textbf{DNA methylation}: DNA methylation is an epigenetic modification involving the addition of methyl groups to the DNA molecule, typically at cytosine bases adjacent to guanine, known as CpG sites \cite{loyfer2023dna}. This modification plays a crucial role in regulating gene expression without altering the DNA sequence \cite{loyfer2023dna}. In cancer, aberrant methylation can lead to the silencing or activation of genes, contributing to oncogenesis and tumor progression \cite{lakshminarasimhan2016role}. Analyzing methylation profiles across different cancer types helps identify risk and diagnostic markers, predict disease progression, and support personalized treatment strategies \cite{lakshminarasimhan2016role}. DNA methylation is quantified through beta values ranging from 0 to 1, with higher values indicating increased methylation \cite{du2010comparison}. The beta values for TCGA-GDC methylation data were obtained using the Illumina Human Methylation 450 platform, which provides detailed methylation profiling \cite{wang2018framework}. The dataset contains 485,576 unique cg and rs methylation sites across multiple tumor types \cite{wang2018framework}.

                \item \textbf{Gene expression (RNAseq)}: Gene expression analysis through RNA sequencing (RNAseq) is a powerful modality in cancer research, providing insights into the transcriptomic landscape of tumors \cite{corchete2020systematic}. This technique quantifies the presence and quantity of RNA in a biological sample, giving a detailed view of transcriptional activity in a cell \cite{corchete2020systematic}. RNAseq helps identify genes that are upregulated or downregulated in cancer cells compared to normal cells, offering clues about oncogenic pathways and potential therapeutic targets \cite{hijazo2021gene}. TCGA-GDC gene expression data was obtained from RNAseq, utilizing High-throughput sequence Fragments Per Kilobase of transcript per Million mapped reads (HTseq-FPKM) for normalization \cite{gonzalez2023gene}. This approach normalizes raw read counts by gene length and the number of mapped reads, with further processing involving incrementing the FPKM value by one followed by log transformation to stabilize variance and enhance statistical analysis \cite{rau2019exploring}. The dataset includes 60,483 genes, with FPKM values indicating gene expression levels. Values above 1000 signify high expression, while values between 0.5 and 10 indicate low expression \cite{gonzalez2023gene,genexpression}.

                \item \textbf{miRNA stem loop expression}:  miRNA stem-loop expression plays a pivotal role in understanding the regulatory mechanisms of miRNAs (microRNAs) in gene expression \cite{peng2016role}. miRNAs are small, non-coding RNA molecules that function by binding to complementary sequences on target mRNA transcripts, leading to silencing \cite{peng2016role}. The expression of miRNAs involves multiple steps to ensure specific targeting and effective modulation of gene expression, which is crucial for normal cellular function as well as pathological conditions like cancer \cite{peng2016role}. miRNA expression values for TCGA-GDC were measured using stem-loop expression through Illumina, and values were log-transformed after the addition of one \cite{chu2016large,lin2022integrative}. The data represents 1880 features across hsa-miRNA sites, with expression levels varying between high and low.

                \item \textbf{Protein expression}:  Reverse Phase Protein Array (RPPA) is a laboratory technique similar to western blotting, used to quantify protein expression in tissue samples \cite{GDCDocs}. The method involves transferring antibodies onto nitrocellulose-coated slides to bind specific proteins, forming quantifiable spots via a DAB calorimetric reaction and tyramide dye deposition, analyzed using "SuperCurve Fitting" software \cite{GDCDocs, RPPAdescription}. RPPA effectively compares protein expression levels in tumor and benign samples, highlighting aberrant protein levels that define the molecular phenotypes of cancer \cite{GDCDocs,chen2019tcpa}. RPPA data in TCGA was derived from profiling nearly 500 antibody-proteins for each patient and deposited in The Cancer Proteome Atlas portal \cite{li2013tcpa}. Each dataset includes the antigen ID, peptide target ID, gene identifier that codes for the protein, and antigen expression levels. Protein expression levels were normalized through log transformation and median centering after being calculated by SuperCurve fitting software \cite{SuperCurve}.
                
                \item \textbf{DNA mutation}: Analyzing DNA sequences involves identifying mutated regions compared to a reference genome, resulting in Variant Calling Format (VCF) files detailing these differences \cite{dnamut3, dnamut1}. Aggregating VCF files to exclude low-quality variants and include only somatic mutations produces Mutation Annotation Format (MAF) files \cite{dnamut2}. Unlike VCF files, which consider all reference transcripts, MAF files focus on the most affected references and include detailed characteristics and quantifiable scores that assess a mutation's translational impact and clinical significance \cite{dnamut2}. This information is critical because clinically significant mutations often result in major defects in protein structure, severely impacting downstream functions and contributing to cancer development \cite{mendiratta2021cancer}. The MAF files from TCGA-GDC contain 18,090 mutational characteristics \cite{dnamut2}.
                
                \item \textbf{Clinical data}:  Clinical and patient-level data play a crucial role in cancer research, providing the foundation for identifying and characterizing patient cohorts \cite{morin2021artificial}. Clinical data includes detailed patient information that is instrumental in understanding cancer epidemiology, evaluating treatment responses, and improving prognostic assessments \cite{morin2021artificial}. Integrating clinical data with genomic and proteomic analyses can uncover relationships between molecular profiles and clinical manifestations of cancer \cite{waqas2023multimodal}. Key clinical and patient-level covariates such as age, gender, race, and disease stage are particularly important in cancer research due to their impact on disease presentation, progression, and treatment efficacy \cite{lewandowska2022risk, lopes2020genome, zavala2021cancer, yang2022research}. Age is a critical factor as cancer incidence and type often vary significantly with age, influencing both the biological behavior of tumors and patient prognosis \cite{lewandowska2022risk}. Gender also plays an important role, with certain cancers being gender-specific and others differing in occurrence and outcomes between genders due to biological, hormonal, and social factors \cite{lopes2020genome}. Race and ethnicity are linked to differences in cancer susceptibility, mortality rates, and treatment outcomes, which reflect underlying genetic, environmental, and socioeconomic factors \cite{zavala2021cancer}. Finally, cancer stage and histology at diagnosis are paramount for determining disease extent, guiding treatment decisions, and correlating directly with survival rates \cite{yang2022research}.

            \end{enumerate}

        \subsubsection{Pre-processing}
        Multiomics data integrates diverse biological data modalities such as genomics, transcriptomics, proteomics, and metabolomics, to understand the complex mechanisms of diseases like cancer. However, before integration, this data requires multiple preprocessing steps to overcome the \emph{big P, small n} problem and other associated challenges of high-throughput molecular data. The \emph{big P, small n} problem refers to a large number of features (P) and a small number of samples (n) in the data \cite{liao2007logistic}. The pan-cancer multi-omics data comes with intra- and inter-dataset correlations, heterogeneous measurement scales, missing values, technical variability, and other background noise. Key challenges include: (i) data heterogeneity, where each data type has unique properties and scales, (ii) volume and complexity, which involve managing and processing overwhelming volumes of data, often in terabytes, (iii) quality and variability, which stem from different platforms causing batch effects, sensitivity differences, noise, varying error rates, and missingness, and (iv) lack of standardization in data collection and processing across laboratories and studies. These challenges complicate the preprocessing needed to make the data machine learning-ready. The key preprocessing tasks for multi-omic data are:

            \begin{enumerate}
                \item \textbf{Normalization and scaling}.  Due to their diverse nature, each omics data type requires specific normalization techniques (e.g., gene length adjustment in RNA-seq or protein abundance correction in proteomics). Choosing the right normalization method ensures that data are comparable across modalities \cite{zhao2021tpm,kaushik2014spatial,liu2014comprehensive}.

                \item \textbf{Handling missing data}.   Multiomics datasets often contain missing values due to detection limits or experimental errors. In some cases, an entire data modality for a patient may be missing. Robust imputation methods are critical to avoid biased interpretations. Common methods include mean, median, kNN, Gaussian mixture clustering, Bayesian approaches, and deep learning-based techniques such as autoencoders \cite{song2020review}.

                \item \textbf{Dimensionality reduction}.    The high dimensionality of multi-omics data often exceeds the number of samples available, increasing the risk of overfitting. Techniques like principal component analysis, t-distributed stochastic neighbor embedding, feature selection, and feature engineering are used to reduce dimensionality while preserving the most informative aspects of the data \cite{anowar2021conceptual}.

                \item \textbf{Data annotation and metadata}.    Proper annotation and comprehensive metadata are essential for effective preprocessing of multiomics data. Metadata should capture details about sample collection, processing protocols, and experimental conditions to ensure accurate data interpretation and reproducibility \cite{settino2018survey}.

                \item \textbf{Integration techniques}.  Integrating diverse datasets involves sophisticated statistical and computational methods. Techniques such as concatenation, transformation, and advanced modeling (e.g., machine learning or deep learning algorithms) are typically used to merge these datasets coherently \cite{lei2023tcga}.
            \end{enumerate}

        Addressing these challenges requires interdisciplinary expertise, including bioinformatics, statistics, and domain-specific knowledge. Here, we describe the preprocessing steps used across molecular data modalities.

\begin{figure}[ht!]%
            \centering
            \includegraphics[width=1.0\textwidth]{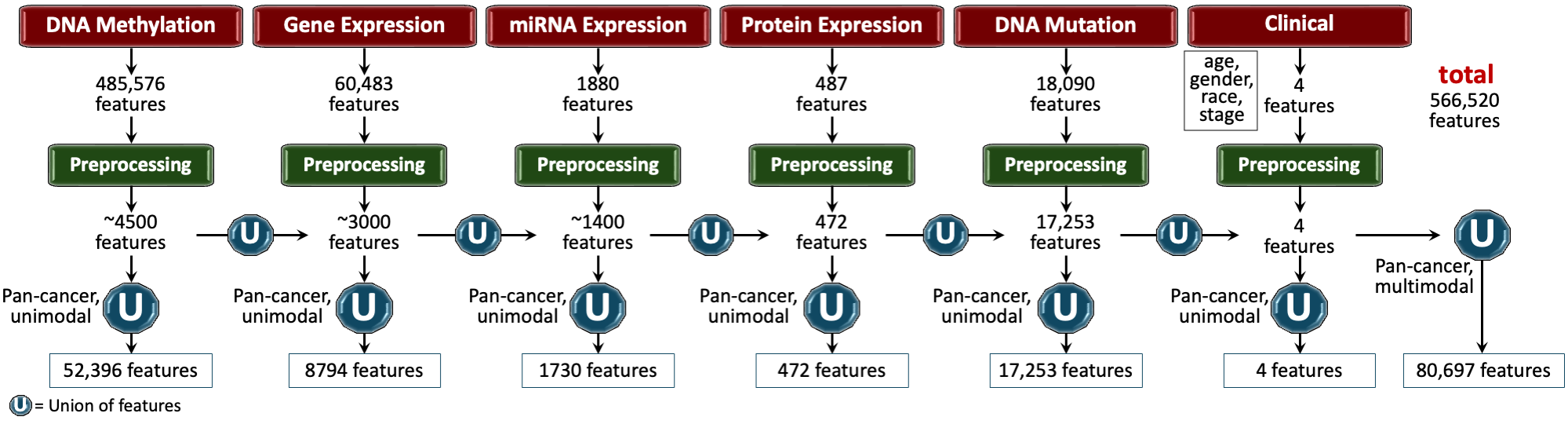}
            \caption{Features processing pipeline for pan-cancer data encompassing six data types: DNA Methylation, Gene Expression, miRNA Expression, Protein Expression, DNA Mutation, and Clinical features. Initial feature counts are reduced through preprocessing, with each modality unified at the pan-cancer level to yield unimodal pan-cancer feature sets. These unimodal features are further unified across all modalities, resulting in a final integrated multimodal feature matrix of 80,697 features, used for downstream analysis.}\label{fig:preprocess}
        \end{figure}

        \begin{itemize}
            \item \textbf{Remove NaNs}.  First, we removed the features that had NaNs across all the samples. This reduced the dimension, removed noise, and ensured continuous-numbered features to work with.
          
            \item \textbf{Drop constant features}.  Next, constant/quasi-constant features with a threshold of 0.998 were filtered out using Feature-engine, a Python library for feature engineering and selection \cite{feture-engine}. This eliminated features with no expression at all across every sample along with features that were noise, since the expression value was the same across every sample. 
            
            \item \textbf{Remove duplicates features}.    Next, duplicate features between genes were identified that contained the same values across two seperate genes, and one of the genes was kept. This may reveal gene-gene relationships between the two genes stemming from an up-regulation pathway or could simply reflect noise. 
            
            \item \textbf{Remove colinear features}.    Next, we filtered the features having low variance ($\approx$0.25) because the features having high variance hold the maximum amount of information \cite{bommert2022benchmark}. We used VarianceThreshold feature selector of scikit learn library that removes low-variance features based on the defined threshold \cite{scikit-learn}. We chose a threshold for each data modality so that the resulting features have matching dimensions, as shown in Figure \ref{fig:preprocess}.

            \item \textbf{Remove low-expression genes}. The gene expression data originally contained 60,483 features, with FPKM transformed numbers ranging from 0 to 12. Roughly 30,000 genes remained after the above-mentioned preprocessing steps, which was still a very high number of features. High expression values reveal important biological insights due to an indication that a certain gene product is transcribed in large quantities, revealing that gene features with large expression values within the dataset are highly relevant. Genes containing an expression value greater than 7 (127 FPKM value) were kept, while the rest were discarded. Around 3,000 genes remained after this process, all of which ranged from values between 7 and 12.

            \item \textbf{Handle missing features}. We handled missing features at two levels of data integration. First, for the features within each modality and cancer type, the missing values were imputed with the mean of the samples for that feature. This resulted in the full-length feature vector for each sample. Second, across different cancers and modalities, we padded the missing features with zeros. One may opine that this is equivalent to zero-padding prevalent in the bio-statistics, but we argue that padding zeros across cancers and modalities is not an imputation when integrating very high dimensional, and high-sample-sized data. In deep learning, the zero imputation technique shows the best performance compared to other imputation techniques and deficient data removal techniques \cite{anggraeny2018analysis, ulriksborg2022imputation}. Moreover, there is a line of work that simply used zero padding to minimize the noise in data and achieved state-of-the-art performance on respective datasets \cite{talwar2018autoimpute, yi2019not}.
        \end{itemize}

        \subsubsection{Features integration}
        After carrying out the preprocessing steps mentioned above, we integrate the data across cancers and across modalities. We generate two views of the data by combining the features across cancers and across modalities. First view is created by taking the union of features across all cancer patients for each of the six modalities (DNA methylation, gene expression, miRNA expression, protein expression, DNA mutation, and clinical). As a result of the preprocessing explained earlier, the DNA methylation data features were reduced from 485,576 features to $\approx$ 4,500 features for all cancers. The union of these features from individual cancers resulted in a feature dimension of 52,396. The gene expression data originally had 60,483 features across all cancers, which was reduced to $\approx$ 3000 features. Union of these features resulted in the feature dimension of 8,794. The miRNA expression data originally had 1,880 features across all cancers, which was reduced to $\approx$ 1,400 features. Union of these features resulted in the feature dimension of 1,730. The protein expression data originally had 487 features across all cancers, which was reduced to 472 features unionized to 472 dimensions. The DNA mutation data had 18,090 features across all cancers, pre-processed and unionized to 17,253 features. Lastly, we convert the categorical clinical features to numerical values such as gender, race, and cancer stages. The details of these clinical characteristics are given in Table \ref{Tab:clinchars}. Mathematically, the preprocessing is given below.
        
        Let \textbf{v} represent the initial feature having fixed dimension for each cancer. The dimension of each feature set is reduced through a preprocessing step, resulting in the feature vector $\tilde{\textbf{v}}$, which is calculated by a function of \textbf{v}, noted as \( f(\textbf{v}) \), where \( f \) is the dimension reduction function such as those presented in the previous section, $\tilde{\textbf{v}} = f(\textbf{v})$. For \( n =33 \) cancer types, the reduced dimensional feature vector $\tilde{\textbf{v}}$ from each cancer type are then combined through a union operation to generate a feature vector \( V_m \) for each modality \( m \) and \( M = 6 \) are the total number of modalities. The feature vector for each modality, \( V_m \), is defined as:
        \begin{equation}
        V_m = 
            \begin{cases} 
            \bigcup_{i=1}^n \tilde{\textbf{v}}_i & \text{if } \tilde{\textbf{v}}_i \text{ varies by cancer type or modality,} \\
            \tilde{\textbf{v}} & \text{otherwise.}
            \end{cases}
        \end{equation}

        Finally, the union of all \( V_m \) across different modalities results in the total pan-cancer, multimodal feature vector \( V_c \in \mathbb{R}^{80,697}\). The total pan-cancer, multimodal feature vector \( V_c \) can then be expressed as:
        \begin{equation}
            V_c = \bigcup_{m=1}^M V_m
        \end{equation}

        \begin{table}[ht]
            \centering
            \caption{Feature Reduction Summary of Pan-cancer data.}   
            \tiny 
            \begin{tabular}{
            >{\raggedright\arraybackslash}p{1.5cm} >{\raggedright\arraybackslash}p{2.4cm}
            >{\raggedright\arraybackslash}p{0.5cm} >{\raggedright\arraybackslash}p{0.5cm} 
            >{\raggedright\arraybackslash}p{0.5cm} >{\raggedright\arraybackslash}p{0.5cm} 
            >{\raggedright\arraybackslash}p{0.5cm} >{\raggedright\arraybackslash}p{0.5cm} 
            >{\raggedright\arraybackslash}p{0.5cm} >{\raggedright\arraybackslash}p{0.5cm} 
            >{\raggedright\arraybackslash}p{0.5cm} >{\raggedright\arraybackslash}p{0.5cm} >{\raggedright\arraybackslash}p{0.5cm}}      
            \toprule
            \textbf{Data} & \textbf{Primary Site} & \textbf{Cases} & 
            \multicolumn{2}{c}{\textbf{miRNA Exprn}} & \multicolumn{2}{c}{\textbf{DNA Methyl}} &
            \multicolumn{2}{c}{\textbf{Gene Exprn}} & \multicolumn{2}{c}{\textbf{Protein Exprn}} &
            \multicolumn{2}{c}{\textbf{DNA Mut}} \\
            \cmidrule(lr){4-5} \cmidrule(lr){6-7} \cmidrule(lr){8-9} \cmidrule(lr){10-11} \cmidrule(lr){12-13}
            &  &  & Before & After & Before & After & Before & After & Before & After & Before & After \\
            \midrule
            TCGA-DLBC   & Large B-cell Lymphoma     & 51 & 1880 &    1060 & 485576 & 4396 & 60483 & 850 & 487 & 472 & 18090 & 17253 \\
            TCGA-UCS    & Uterine Carcinosarcoma    & 61 & 1880 &    1101 & 485576 & 4632 & 60483 & 1231 & 487 & 472 & 18090 & 17253 \\
            TCGA-CHOL   & Bile Duct                 & 62 & 1880 &    967  & 485576 & 4479 & 60483 & 1261 & 487 & 472 & 18090 & 17253 \\
            TCGA-UVM    & Uveal melanomas           & 80 & 1880 &    1162 & 485576 & 4019 & 60483 & 772 & 487 & 472 & 18090 & 17253 \\
            TCGA-MESO	& Mesothelioma	            & 86 & 1880	&    1158 & 485576 & 4372 & 60483 & 1278 & 487 & 472 & 18090 & 17253 \\
            TCGA-ACC	& Adrenocortical	        & 95 &	1880&    1110 & 485576 & 4454 & 60483 & 1304 & 487 & 472 & 18090 & 17253 \\
            TCGA-THYM	& Thymoma	                & 138&	1880&    1245 & 485576 & 4609 & 60483 & 1337 & 487 & 472 & 18090 & 17253 \\
            TCGA-TGCT	& Testicular	            & 139&	1880&    1290 & 485576 & 4762 & 60483 & 1343 & 487 & 472 & 18090 & 17253 \\
            TCGA-READ	& Rectal	                & 178&	1880&    1314 & 485576 & 4077 & 60483 & 1547 & 487 & 472 & 18090 & 17253 \\
            TCGA-KICH	& Kidney Chromophobe	    & 182&	1880&    1089 & 485576 & 4333 & 60483 & 1107 & 487 & 472 & 18090 & 17253 \\
            TCGA-PCPG	& Pheochromocytoma and Paraganglioma& 189&	1880&1251&485576&4550 & 60483 & 1216 & 487 & 472 & 18090 & 17253 \\            
            TCGA-PAAD	& Pancreatic	            & 222&	1880&    1308 & 485576 & 4518 & 60483 & 1567 & 487 & 472 & 18090 & 17253 \\
            TCGA-ESCA	& Esophageal	            & 249&	1880&    1300 & 485576 & 4192 & 60483 & 1684 & 487 & 472 & 18090 & 17253 \\
            TCGA-SARC	& Sarcoma	                & 287&	1880&    1235 & 485576 & 4467 & 60483 & 2490 & 487 & 472 & 18090 & 17253 \\
            TCGA-CESC	& Cervical	                & 304&	1880&    1405 & 485576 & 4167 & 60483 & 2017 & 487 & 472 & 18090 & 17253 \\
            TCGA-KIRP	& Kidney Papillary Cell Carcinoma&376&1880&  1297 & 485576 & 4078 & 60483 & 1798 & 487 & 472 & 18090 & 17253 \\
            TCGA-SKCM	& Skin Cutaneous Melanoma	& 436&	1880&    1426 & 485576 & 4427 & 60483 & 2488 & 487 & 472 & 18090 & 17253 \\
            TCGA-BLCA	& Bladder	                & 447&	1880&    1361 & 485576 & 4483 & 60483 & 2751 & 487 & 472 & 18090 & 17253 \\
            TCGA-LIHC	& Liver	                    & 463&	1880&    1336 & 485576 & 4023 & 60483 & 2017 & 487 & 472 & 18090 & 17253 \\
            TCGA-STAD	& Stomach	                & 499&	1880&    1397 & 485576 & 4196 & 60483 & 2354 & 487 & 472 & 18090 & 17253 \\
            TCGA-LGG	& Lower Grade Glioma	    & 533&	1880&    1287 & 485576 & 4193 & 60483 & 1560 & 487 & 472 & 18090 & 17253 \\
            TCGA-COAD	& Colon	                    & 539&	1880&    1460 & 485576 & 4671 & 60483 & 1931 & 487 & 472 & 18090 & 17253 \\
            TCGA-UCEC	& Endometrioid	            & 588&	1880&    1414 & 485576 & 4424 & 60483 & 2849 & 487 & 472 & 18090 & 17253 \\
            TCGA-HNSC	& Head and Neck	            & 611&	1880&    1428 & 485576 & 4358 & 60483 & 2059 & 487 & 472 & 18090 & 17253 \\
            TCGA-THCA	& Thyroid	                & 614&	1880&    1369 & 485576 & 4160 & 60483 & 1432 & 487 & 472 & 18090 & 17253 \\
            TCGA-PRAD	& Prostate	                & 623&	1880&    1334 & 485576 & 4006 & 60483 & 1635 & 487 & 472 & 18090 & 17253 \\
            TCGA-LAML	& Acute Myeloid Leukemia	& 626&	1880&    1140 & 485576 & 4415 & 60483 & 1032 & 487 & 472 & 18090 & 17253 \\
            TCGA-GBM	& Glioblastoma	            & 649&	1880&    1023 & 485576 & 4076 & 60483 & 1206 & 487 & 472 & 18090 & 17253 \\
            TCGA-LUAD	& Lung Adenocarcinoma	    & 728&	1880&    1360 & 485576 & 4480 & 60483 & 2562 & 487 & 472 & 18090 & 17253 \\
            TCGA-OV	    & Ovarian	                & 731&	1880&    1430 & 485576 & 4254 & 60483 & 2116 & 487 & 472 & 18090 & 17253 \\            
            TCGA-LUSC	& Lung Squamous Cell Carcinoma& 752&1880&    1375 & 485576 & 4302 & 60483 & 2610 & 487 & 472 & 18090 & 17253 \\
            TCGA-KIRC	& Kidney Clear Cell Carcinoma& 979&	1880&    1333 & 485576 & 4399 & 60483 & 2274 & 487 & 472 & 18090 & 17253 \\
            TCGA-BRCA	& Breast	               & 1260&	1880&    1418 & 485576 & 4195 & 60483 & 3671 & 487 & 472 & 18090 & 17253 \\
           \bottomrule
            \end{tabular} \label{Tab:features}
            \end{table}
            
            
        \begin{table}[ht]
            \centering
            \caption{Summary of patient characteristics for pan-cancer data used in this study.}   
            \scriptsize 
            \begin{tabular}{
            >{\raggedright\arraybackslash}p{1.5cm}
            >{\raggedright\arraybackslash}p{1.42cm} >{\raggedright\arraybackslash}p{0.6cm} 
            >{\raggedright\arraybackslash}p{3.0cm} >{\raggedright\arraybackslash}p{5.4cm} 
            }            
            \toprule
            \textbf{Cancer Type} & \textbf{Age (Mean±SD)} & 
            \textbf{Gender (M/F)} & \textbf{Race} (White/Asian/Black/NA/American Indian/Alaska/Islander) &
            \textbf{Stage} (0/I/IA/IB/IC/II/IIA/IIB/IIC/III/IIIA/IIIB/IIIC/IV/IVA/IVB/IVC/NA) \\
            
            \midrule
            TCGA-ACC    &   47.46 ± 16.2    &   33/62   &   79/3/1/12/0/0       &   0/9/0/0/0/46/0/0/0/20/0/0/0/17/0/0/0/3      \\
            TCGA-BLCA   &   67.92 ± 10.39   &   326/121 &   363/43/23/18/0/0    &   0/3/0/0/0/136/0/0/0/159/0/0/0/148/0/0/0/1   \\
            TCGA-BRCA   &   57.94 ± 13.11    &   13/1247 &   915/59/198/87/1/0   &   0/114/94/7/0/6/404/307/0/2/176/30/74/22/0/0/0/24 \\
            TCGA-CESC   &   48.04 ± 13.7     &   0/304   &   211/19/32/30/9/0    &   0/0/0/0/0/0/0/0/0/0/0/0/0/0/0/0/0/304       \\
            TCGA-CHOL   &   64.37 ± 12.21    &   30/32   &   55/3/3/1/0/0        &   0/30/0/0/0/16/0/0/0/5/0/0/0/2/3/6/0/0       \\
            TCGA-COAD   &   66.93 ± 12.67    &   288/251 &   261/11/67/198/2/0   &   0/87/1/0/0/46/150/13/2/26/9/69/47/56/18/3/0/12  \\
            TCGA-DLBC   &   56.76 ± 13.68    &   24/27   &   32/18/1/0/0/0       &   0/0/0/0/0/0/0/0/0/0/0/0/0/0/0/0/0/51        \\
            TCGA-ESCA   &   64.22 ± 12.11    &   208/41  &   162/46/6/35/0/0     &   0/14/9/7/0/1/56/43/0/41/16/10/9/7/6/0/0/30  \\
            TCGA-GBM    &   57.74 ± 14.32    &   399/250 &   547/13/53/36/0/0    &   0/0/0/0/0/0/0/0/0/0/0/0/0/0/0/0/0/649       \\
            TCGA-HNSC   &   61.02 ± 11.92    &   443/168 &   522/12/58/17/2/0    &   0/29/0/0/0/93/0/0/0/97/0/0/0/0/302/13/1/76  \\
            TCGA-KICH   &   51.61 ± 14.12    &   99/83   &   154/6/19/3/0/0      &   0/75/0/0/0/59/0/0/0/34/0/0/0/14/0/0/0/0     \\
            TCGA-KIRC   &   60.67 ± 11.95    &   641/338 &   876/16/73/14/0/0    &   0/475/0/0/0/102/0/0/0/237/0/0/0/161/0/0/0/4 \\
            TCGA-KIRP   &   61.98 ± 12.2     &   278/98  &   275/6/75/16/4/0     &   0/219/0/0/0/25/0/0/0/77/0/0/0/21/0/0/0/34   \\
            TCGA-LAML   &   54.82 ± 15.87    &   345/281 &   564/8/49/5/0/0      &   0/0/0/0/0/0/0/0/0/0/0/0/0/0/0/0/626         \\
            TCGA-LGG    &   42.71 ± 13.32    &   293/240 &   492/8/22/10/1/0     &   0/0/0/0/0/0/0/0/0/0/0/0/0/0/0/0/533         \\
            TCGA-LIHC   &   60.44 ± 13.71    &   305/158 &   255/168/25/14/1/0   &   0/211/0/0/0/105/0/0/0/6/78/12/11/2/1/3/0/34 \\
            TCGA-LUAD   &   65.20 ± 10.08     &   329/399 &   580/14/84/48/2/0    &   0/7/194/195/0/2/67/103/0/0/101/12/0/37/0/0/0/10 \\
            TCGA-LUSC   &   67.28 ± 8.62     &   548/204 &   530/12/47/163/0/0   &   0/4/127/243/0/4/87/138/0/3/94/33/0/12/0/0/0/7   \\
            TCGA-MESO   &   63.01 ± 9.78     &   70/16   &   84/1/1/0/0/0        &   0/7/2/1/0/15/0/0/0/45/0/0/0/16/0/0/0/0      \\
            TCGA-OV     &   59.60 ± 11.44     &   0/731   &   626/25/43/33/3/0    &   0/0/0/0/0/0/0/0/0/0/0/0/0/0/0/0/731         \\
            TCGA-PAAD   &   64.87 ± 11.36    &   123/99  &   195/13/8/6/0/0      &   0/1/6/15/0/0/36/148/0/6/0/0/0/7/0/0/0/3     \\
            TCGA-PCPG   &   47.02 ± 15.15    &   84/105  &   157/7/20/4/1/0      &   0/0/0/0/0/0/0/0/0/0/0/0/0/0/0/0/189         \\
            TCGA-PRAD   &   60.93 ± 6.8      &   623/0   &   510/13/81/18/1/0    &   0/0/0/0/0/0/0/0/0/0/0/0/0/0/0/0/623         \\
            TCGA-READ   &   63.83 ± 11.85    &   98/80   &   90/1/7/80/0/0       &   0/37/0/0/0/7/40/2/1/6/7/25/14/21/7/0/0/11   \\
            TCGA-SARC   &   60.70 ± 14.38    &   129/158 &   253/5/20/9/0/0      &   0/0/0/0/0/0/0/0/0/0/0/0/0/0/0/0/287         \\
            TCGA-SKCM   &   57.84 ± 15.41    &   289/174 &   441/12/1/9/0/0      &   6/30/18/30/0/39/18/28/61/44/16/46/68/23/0/0/0/36    \\
            TCGA-STAD   &   65.44 ± 10.53    &   320/179 &   311/108/15/64/0/0   &   0/1/21/46/0/37/54/71/0/4/88/67/39/47/0/0/0/24   \\
            TCGA-TGCT   &   31.87 ± 9.19     &   139/0   &   124/4/6/5/0/0       &   0/69/26/11/0/4/6/1/1/2/1/6/5/0/0/0/0/7      \\
            TCGA-THCA   &   47.17 ± 15.83    &   166/448 &   413/59/35/106/1/0   &   0/350/0/0/0/64/0/0/0/134/0/0/0/4/52/0/8/2   \\
            TCGA-THYM   &   58.12 ± 13       &   72/66   &   115/13/8/2/0/0      &   0/0/0/0/0/0/0/0/0/0/0/0/0/0/0/0/138         \\
            TCGA-UCEC  &   63.74 ± 11.06    &   0/588   &   402/21/120/32/4/0   &   0/0/0/0/0/0/0/0/0/0/0/0/0/0/0/0/588         \\
            TCGA-UCS    &   70.07 ± 9.24     &   0/61    &   50/1/9/1/0/0        &   0/0/0/0/0/0/0/0/0/0/0/0/0/0/0/0/61          \\
            TCGA-UVM    &   61.65 ± 13.95    &   45/35   &   55/0/0/25/0/0       &   0/0/0/0/0/0/12/27/0/0/25/10/1/4/0/0/0/1     \\
            Moffitt-LSCC   &	69.14 ± 8.34     &    72/36  &   105/0/3/0/0/0       &   0/0/24/25/0/0/31/15/0/0/12/1/0/0/0/0/0/0    \\
           \bottomrule
            \end{tabular} \label{Tab:clinchars}
            \end{table}

    \subsection{Clinical end-points}
    To assess the performance of the SeNMo framework, we selected three clinical end-points that fall under two categories of machine learning tasks. The first end-point is Overall Survival (OS), which is treated as a regression task. The second is the prediction of primary cancer type, formulated as a 33-class classification task. The third end-point is TLS ratio prediction, also a regression task.            
    
        \subsubsection{Overall Survival (OS)}
\begin{figure}[H]%
            \centering
            \includegraphics[scale=0.5]{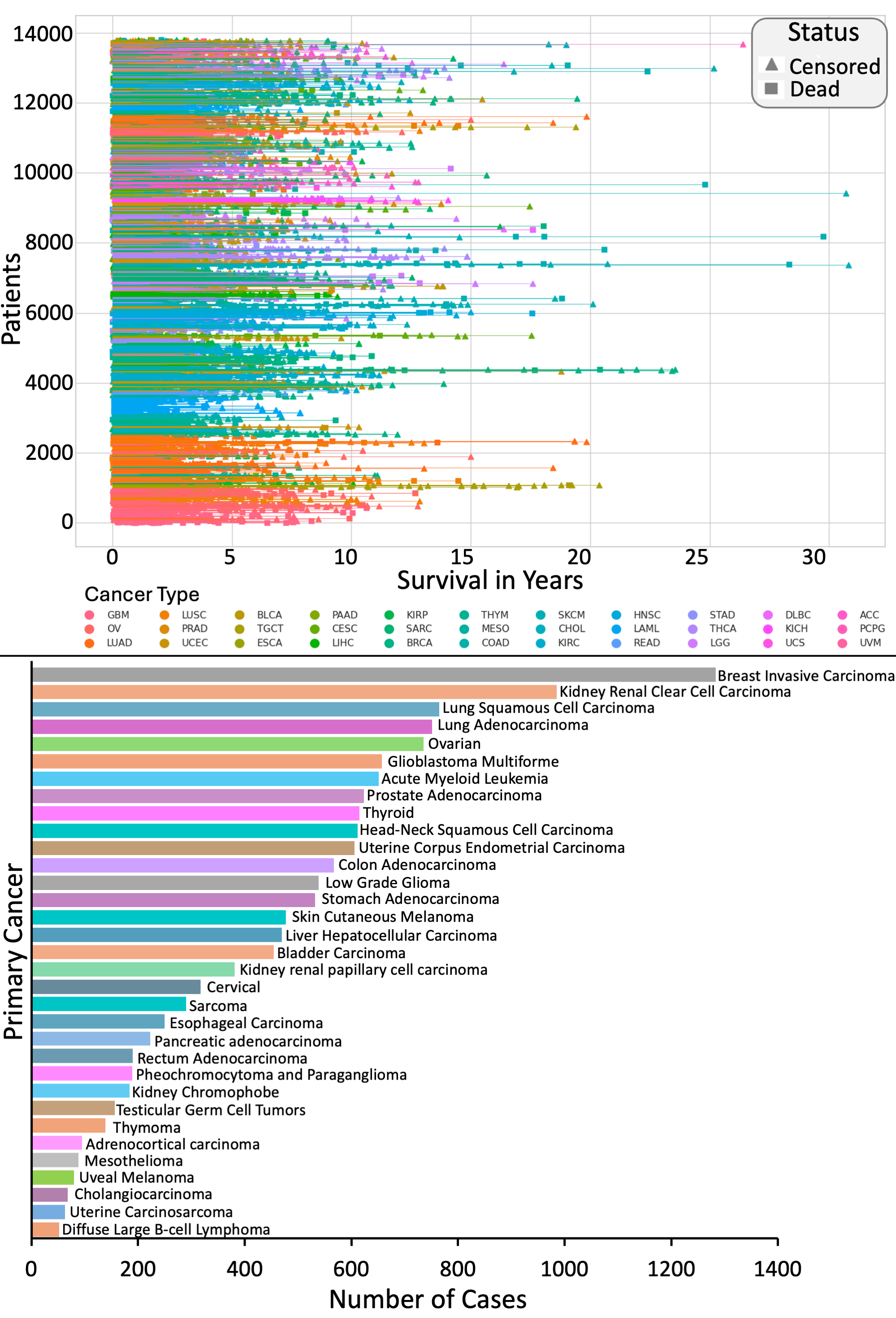} %
            \caption{Summary of the survival data and number of cases in the pan-cancer data. The top panel illustrates patient survival in years, with each line representing an individual patient, marked as censored or dead, and grouped by cancer type. The bottom panel displays the number of cases for each primary cancer type, highlighting variation in sample sizes across cancers, which range from high counts in breast, kidney, and lung to lower counts in rare cancers like diffuse large B-cell lymphoma.}\label{fig:cases}
        \end{figure}

        Predicting cancer prognosis through survival outcomes is a standard approach for biomarker discovery, patient stratification, and assessing therapeutic response \cite{chen2020pathomic}. Statistical survival models, coupled with the integration of deep learning in survival analysis, have significantly advanced the prediction of OS. Prior studies have combined different molecular data types and employed a range of statistical and machine learning methods to predict OS across various datasets \cite{patwardhan2024towards, ma2019integrate, ellen2023autoencoder, nikolaou2024quantifying}. This ongoing effort aims to integrate multiple data types to elucidate the relationship between molecular characteristics and patient outcomes, ultimately achieving more precise prognostic assessments and personalized treatment strategies. In this study, we utilize clinical, demographic, genomic, and other molecular data to explore potential risk factors for cancer patients and to analyze their correlation with the patients' time-to-event, specifically OS. The prediction of OS is implemented as a regression task, with the goal of predicting survival time in days. Time-to-event or survival data records not only the occurrence of events such as death but also the duration from the beginning of the study until the event occurs, or until the patient is lost to follow-up (right censoring). Survival times since cancer diagnosis for the pan-cancer dataset are depicted in Figure \ref{fig:cases}A. Because of censoring, exact survival times are unknown for some patients. In these cases, each patient's outcome is characterized by two variables: a censoring indicator, also known as the vital status, and the observed time $T=\min(T_{s}, T_{\delta})$, where ${T}_s$ represents the true survival time and ${T}_\delta$ is the censoring time, $\{{T}_s \leq {T}_\delta\}$ \cite{zhao2024tutorial}. The survival function, which describes the probability that a patient will survive beyond a specified time $t$, is given by:

            \begin{equation}
                F(t)=P\{T>t\}
            \end{equation}
        
        Additionally, the hazard function provides insight into the risk of an event occurring at a particular time, given survival up to that point. It represents the instantaneous rate of events (e.g., death) occurring at a specific time, conditional on having survived to that time. The hazard function $h(t)$ is mathematically defined as the ratio of the probability of the event occurring in a short interval around $t$ to the probability of surviving beyond $t$:

        \begin{equation}
                h(t) = \lim_{\Delta t \to 0} \frac{P(t \leq T < t + \Delta t \,|\, T \geq t)}{\Delta t},
            \end{equation}

        where, $h(t)$ is the hazard function at time $t$,  $T$ is the survival time, $P(t\leq T< t+\Delta{t} \,|\, T\geq{t})$ is the conditional probability that the event occurs in the time interval $[t,t+\Delta{t})$ given that survival time is greater than or equal to $t$,  and $\Delta{t}$ represents an infinitesimally small time interval. Based on survival data, the hazard function describes the instantaneous risk of experiencing the event of interest at any given time. In our study, right-censoring was defined as censor $\delta=1$ in case of an event (e.g., death), and $0$ otherwise. 

        \subsubsection{Primary Cancer type}
        The prediction of the primary cancer type involves classifying each cancer sample into one of 33 possible cancer types based on biological and clinical features. This classification task is crucial for clinical decision-making, as accurate identification of the primary cancer type is essential for determining the most effective treatment approach, thereby improving patient outcomes and enabling personalized therapies \cite{miller2019cancer}. Cancer treatments and prognoses differ significantly across cancer types, often necessitating specific, tailored interventions that align with the distinct biological characteristics of each type. Correct identification of the primary cancer type also aids in follow-up care and surveillance, increasing the likelihood of early detection of recurrence. Therefore, achieving high accuracy in this classification not only enhances clinical decision-making but also positively impacts patient survival and quality of life. The pan-cancer dataset encompassing 33 cancer types, along with the distribution of patient samples, is depicted in Figure~\ref{fig:cases}.

        \subsubsection{Tertiary Lymphoid Structures (TLS) Ratio} 
        TLSs are organized accumulations of immune cells that resemble secondary lymphoid organs and form in inflamed peripheral tissues, including within cancers \cite{van2024TLS, chen2024TLS}. TLS presence is linked to improved survival rates and favorable responses to immunotherapy across various solid tumors, making TLS quantification a promising predictive and prognostic biomarker \cite{van2024TLS, chen2024TLS}. The TLS ratio, defined as the segmented TLS area over the total tissue area, is correlated with positive immunotherapy outcomes and overall patient prognosis. Recent studies have demonstrated the value of TLS quantification in cancer, highlighting its role in improving clinical decision-making and developing automated TLS segmentation models with high accuracy in multiple cancers \cite{van2024TLS, chen2024TLS}. In this study, we used whole slide images of H\&E and CD20-stained sections imported into Visiopharm software version 2022.03. Visiopharm’s Tissuealign tool was used to co-register serial H\&E and CD20 images for each patient. Using the H\&E image, manually drawn regions of interest (ROIs) were created to segment the tumor and non-tumor regions in each image set. TLSs were detected through a thresholding algorithm, followed by manual review and feature extraction for analysis by an experienced image analysis technician under the guidance of the study pathologist.

        \begin{figure}[ht!]%
            \centering
            \includegraphics[width=1.0\textwidth]{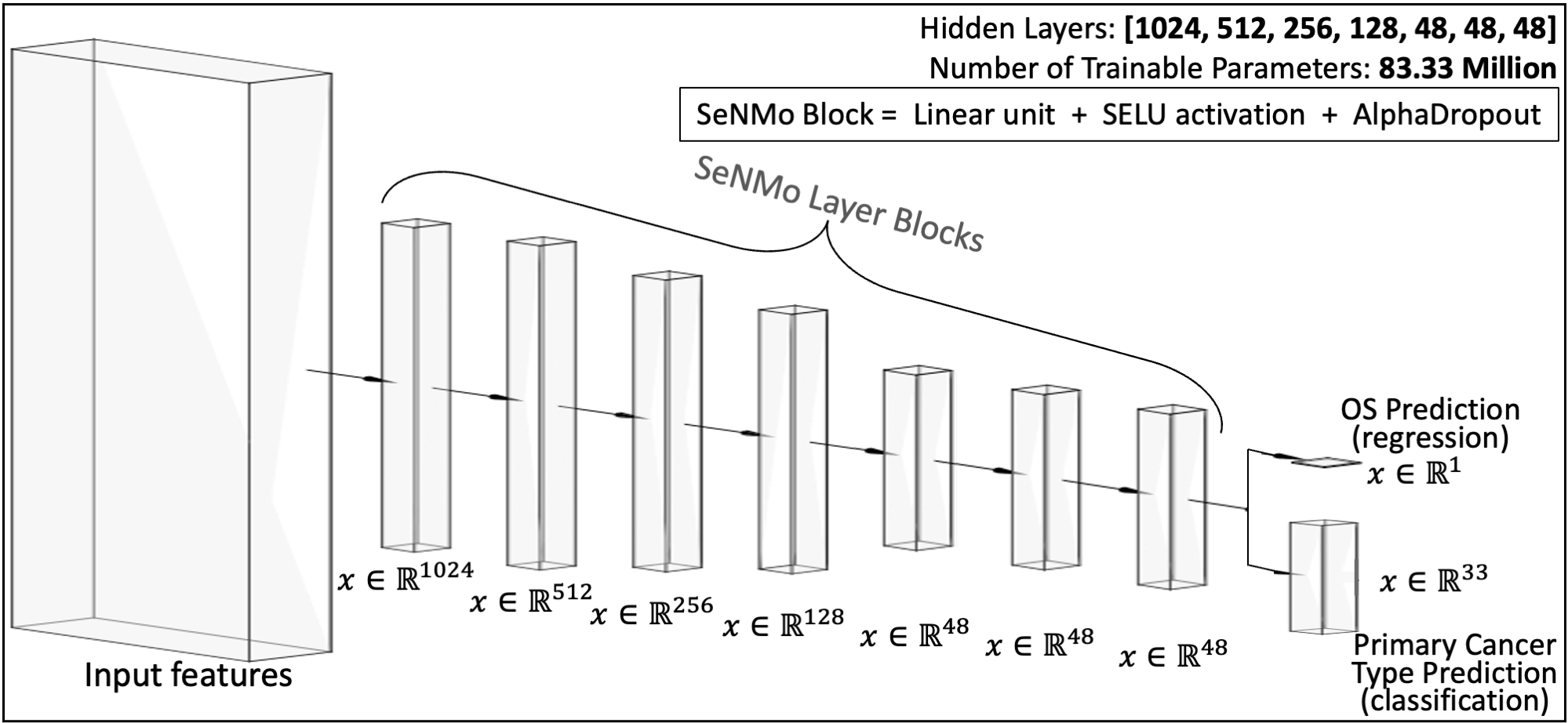} %
            \caption{Architecture of the SeNMo encoder network. There are seven hidden layers each comprising of a linear unit, SELU activation, and alpha-dropout. The trained model has 83.33 million parameters. The number of neurons in each hidden layer, input layer, and output layer are also depicted in the figure. The same model is used for regression and classification tasks.}\label{fig:SeNMoLayers}
        \end{figure}
        
    \subsection{SeNMo Deep Learning Model}
    In scenarios involving hundreds or thousands of features with relatively few training samples, feedforward networks often face the risk of overfitting \cite{chen2020pathomic}. Unlike CNNs, weights in feedforward networks are shared, making them vulnerable to training instabilities caused by perturbations and regularization techniques such as stochastic gradient descent and dropout. CNNs, on the other hand, struggle to handle high-dimensional, low-sample data due to the spatial invariance assumption, fixed input size, and inefficiencies in managing multi-omics data sparsity. Transformer-based models are also suboptimal for high-dimensional, low-sample data, as they rely heavily on attention mechanisms tailored for predicting sequential patterns, which fails when dealing with highly sparse molecular data.

    To address the challenges of overfitting and instability in high-dimensional, low-sample-size multi-omics data, we drew inspiration from self-normalizing networks introduced by Klambauer \textit{et al}. \cite{klambauer2017self}. Self-normalizing neural networks are particularly suited for high-dimensional datasets with limited samples, a characteristic that makes them highly relevant for multi-omics analysis. The SeNMo architecture is based on stacked layers of self-normalizing neural networks, as detailed below.
    
        As illustrated in Figure \ref{fig:SeNMoLayers}, SeNMo comprises stacked blocks of self-normalizing neural network layers, where each block includes a linear unit, a Scaled Exponential Linear Unit (SELU) activation, and Alpha-Dropout. These components enable high-level abstract representations while keeping neuron activations close to zero mean and unit variance \cite{klambauer2017self}. The linear unit is equivalent to a "fully connected" or MLP layer commonly used in traditional neural network architectures. Klambauer \textit{et al}. demonstrated through the Banach fixed-point theorem that activations with close proximity to zero mean and unit variance, propagating through numerous network layers, will ultimately converge to zero mean and unit variance \cite{klambauer2017self}. SELU activations, an alternative to traditional rectified linear unit activations, offer a self-normalizing effect, ensuring activations converge to zero mean and unit variance regardless of the input distribution. The SELU activation function is expressed mathematically as:
        
        \begin{equation}
                \text{SELU}(x) = \lambda 
            \begin{cases}
                x & \text{if } x > 0 \\
                \alpha (e^{x} - 1) & \text{if } x \leq 0
            \end{cases}
        \end{equation}

         where, $\lambda$ is a scaling factor (typically set to $1.05071$) and $\alpha$ is the negative scale factor (typically set to $1.6733$).
        
        Dropout, a regularization method that randomly sets a fraction of input units to zero during training, prevents overfitting. Alpha-Dropout, a modified version of traditional dropout, is designed to maintain the self-normalizing property of SELU activations. It applies a dropout mask during training, scaled to ensure the mean and variance of activations remain stable. The scaling factor is computed based on the dropout rate and the SELU parameters ($\lambda$ and $\alpha$). Alpha-Dropout is mathematically defined as:
        
        \begin{equation}
            \text{Alpha-dropout}(x) = \frac{x - \mu(x)}{std(x)} \times mask + \mu(x)
        \end{equation}
        
        where, $x$ is the input activation,  $\mu(x)$, $std(x)$ are mean and standard deviation of the input activation, respectively, and $mask$ is a binary mask generated with the specified dropout rate.

        Together, SELU activations and Alpha-Dropout ensure that SeNMo blocks maintain stable mean and variance across network layers, facilitating more reliable training and better generalization performance. Additionally, these mechanisms help mitigate training instabilities related to vanishing or exploding gradients in feedforward networks. Our network architecture consists of seven fully connected hidden layers, each followed by SELU activation and Alpha-Dropout, as illustrated in Figure \ref{fig:SeNMoLayers}. The number of neurons in each block is shown in the inset of Figure \ref{fig:SeNMoLayers}. The final fully connected layer is used to learn a latent representation of each sample, termed as the patient embedding $\textbf{x} \in \mathbb{R}^{48}$.   
        
    \subsection{Training and Evaluation}
        \subsubsection{Data Splits}
        For the OS task, the pan-cancer data was randomly divided into the training-validation set ($80\%$) and the hold-out test set ($20\%$) for each cancer type. The pan-cancer training was carried out by combining the training-validation cohort of all $33$ cancer types and adopting the 10-fold cross-validation with the $80-20\%$ division of samples. The training-validation cohort has $11,050$ patients, each having $\mathbb{R}^{80,697}$ features, comprising the six multi-omics modalities, gene expression, DNA methylation, miRNA expression, protein expression, DNA mutation, and the four clinical features (age, gender, race, stage). The SeNMo encoder model was trained on the training-validation cohort for the regression task of predicting the OS. C-Index was used as the evaluation metric of the hazard score predicted by the model. We used weights and biases to find the optimal set of hyperparameters for our deep learning model \cite{wandb}. For the evaluation/testing of the trained model, the inference data was created by combining the held-out test set from all $33$ cancer types, resulting in $2,754$ patients, each having $\mathbb{R}^{80,697}$ features.  We further tested the optimal hyperparameters of our trained model to train different combinations of the pan-cancer data modalities. We call these 1-modal, 3-modal (gene expression, DNA methylation, miRNA expression), 4-modal ($3+$protein expression), 5-modal ($4+$DNA mutation), and 6-modal (all modalities) cohorts. Although our initial model was trained on all 6 modalities, these experiments aim to see how the model performs on each of these pan-cancer cohorts where one or more of the data modalities is missing. 
        
        \subsubsection{Evaluation}
        We evaluate SeNMo's performance with the quantitative and statistical metrics common for survival outcome prediction and classification. For survival analysis, we evaluated the model using the C-index. For the primary cancer type classification, we generate the classification report comprising average accuracy, average precision, recall, F1-score, confusion matrix, and scatter plot. For the TLS Ratio, we employed Huber Loss. We utilized the log-rank test to determine if the survival predictions were statistically significantly different. Below, we explain the loss, evaluation metrics, and statistical tests in detail. 

            \begin{enumerate}            
                \item \textbf{Loss Function}: The loss being used for backpropagation in the model is a combination of three components: Cox loss, cross-entropy loss, and regularization loss. This combined loss function aims to simultaneously optimize the model's ability to predict survival outcomes (Cox loss), encourage model-simplicity or sparsity (regularization loss), and model the likelihood of cancer types (cross-entropy loss). The overall loss is a weighted sum of these three components, where each component is multiplied by a corresponding regularization hyperparameter ($\lambda_{c}$, $\lambda_{ce}$, $\lambda_{r}$). This weighted sum allows for balancing the influence of each loss component on the optimization process. Mathematically, the overall loss can be expressed as:
                        
                        \begin{equation}
                            \text{L} = \text{$\lambda_{c}$} \text{$L_{cox}$} + \text{$\lambda_{ce}$} \text{$L_{ce}$} + \text{$\lambda_{r}$} \text{$L_{reg}$}
                        \end{equation}

                    \begin{itemize}
                        \item Cox proportional hazards loss ($L_{cox}$): Cox loss is a measure of dissimilarity between the predicted hazard scores and the true event times in survival analysis. It is calculated using the Cox proportional hazards model and penalizes deviations between predicted and observed survival outcomes of all individuals who are at risk at time $t_{i}$, weighted by the censoring indicator \cite{Cox-nnet}. The function takes a vector of survival times for each individual in the batch, censoring status for each individual (1 if the event occurred, 0 if censored), and the predicted log hazard ratio for each individual from the neural network, and returns the Cox loss for the batch, which is used to train the neural network via backpropagation. This backpropagation encourages the model to assign higher hazards to high-risk individuals and lower predicted hazards to censored individuals or those who experience the event later. Mathematically, the Cox loss is expressed as:

                                \begin{equation}
                                    \text{$L_{cox}$} = -\frac{1}{N} \sum_{i=1}^{N} \left( \theta_i - \log \sum_{j=1}^{N} e^{\theta_j} \cdot R_{ij} \right) \cdot \delta_i,
                                \end{equation}
                        
                                where $N$ is the batch size (number of samples), $\theta_{i}$ is the predicted hazard for sample $i$, $R_{ij}$ is the indicator function that equals 1 if the survival time of sample $j$ is greater than or equal to the survival time of sample $i$, and 0 otherwise, and $\delta_{i}$ is the censoring indicator for sample $i$, which equals 1 if the event is observed for sample $i$ and 0 otherwise.
        

                        \item Cross-entropy loss ($L_{ce}$): The cross-entropy loss is a common loss function used for multi-class classification problems, particularly when each sample belongs to one of the $C$ classes. When combined with a LogSoftmax layer, the function measures how well a model's predicted log probabilities match the true distribution across various classes. For a multi-class classification problem having $C$ classes, the model's outputs (raw class scores or logits) are transformed into log probabilities using a LogSoftmax layer. The cross-entropy loss compares these log probabilities to the true distribution, which is usually represented in a one-hot encoded format. The loss is calculated by negating the log probability of the true class across all samples in a batch and then averaging these values. For the given output of LogSoftmax, $\log(p_{n,c})$ for each class $c$ in each sample $n$, the cross-entropy loss for a multi-class problem can be defined as:

                        \begin{equation}
                                \text{$L_{ce}$} = -\frac{1}{N} \sum_{n=1}^N \sum_{c=1}^C y_{n,c} \log(p_{n,c}),    
                        \end{equation}

                        where $N$ is the total number of samples, $C$ are the total classes, and $y_{n,c}$ is the target label for sample $n$ and class $c$, typically 1 for the true class and 0 otherwise.
                                                
                        \item Regularization loss ($L_{reg}$): The regularization loss encourages the model's weights to remain small or sparse, thus preventing overfitting and improving generalization. We used $L1$ regularization to the SeNMo's parameters, which penalizes the absolute values of the weights.
                    \end{itemize}

                \item \textbf{Concordance Index (C-index)}: The C-index is a frequently used evaluation metric in survival analysis to assess the predictive accuracy of a model for the time-to-event outcomes \cite{zhao2024tutorial}. It measures the degree to which the model's predictions correlate with the actual survival times observed in the data. It quantifies the model's ability to correctly rank pairs of subjects based on their predicted survival times. The C-index evaluates the probability that, in a randomly selected pair of individuals, the one who experienced the event (like death or failure) first also had a higher risk score predicted by the model. Risk score is the output of the survival model and represents the expected order of the events; the higher the score, the higher the risk of experiencing the event sooner \cite{zhao2024tutorial}. We used the $concordance\_index$ Lifelines function to calculate the C-index \cite{Lifelines}. This function takes the predicted hazard scores for each individual, the true event indicator (e.g., 1 if an event occurred, 0 if censored) for each individual, and the survival times (time to event or censoring) for each individual. The C-index function computes the fraction of all pairs of subjects whose predicted event times are correctly ordered among all pairs where one subject experienced an event and the other did not. C-index ranges between 0 and 1 where 0.5 is the expected result from random predictions, 1.0 is a perfect concordance, and 0.0 is perfect anti-concordance \cite{zhao2024tutorial}. Mathematically,
                    
                    \begin{equation}
                        \begin{split}
                            \text{C-Index} &= \frac{(\text{Number of concordant pairs} + 0.5 \times \text{tied pairs})}{\text{Total number of evaluable pairs}}, \\
                            \text{C-Index} &= \Pr(\hat{S}_i < \hat{S}_j | T_i < T_j, \delta_i = 1) 
                        \end{split} \label{eq:9}
                    \end{equation}
                    where concordant pairs are pairs of individuals where the predicted survival times are correctly ordered relative to the observed survival times, tied pairs are the number of pairs where the predictions are equal or survival times are the same. Total number of evaluable pairs are the total pairs considered, excluding pairs with censoring issues or other exclusions, $\hat{S}_i$ and $\hat{S}_j$ represent the predicted risks or survival probabilities for individuals $i$ and $j$, respectively. $T_i < T_j$ implies that individual $i$ experienced the event before individual $j$, and $\delta_i=1$ indicates that the event for individual $i$ was observed (not censored).
                
                \item \textbf{Cox log-rank function}: The Cox log-rank function calculates the p-value using the log-rank test based on predicted hazard scores, censor values, and the true OS times. The log-rank test is a statistical method to compare the survival distributions of two groups or more groups, where the null hypothesis is that there is no difference between the groups. It is commonly used in survival analysis to compare the observed number of events in each group to the number of events expected under the null hypothesis. For the hazard ratio $h_{i}(t)$ of group $i$ at time $t$, the hypotheses are given by,
                    \begin{equation}
                        \begin{split}
                             & H_0: h_1(t) = h_2(t) \\
                             & H_A: h_1(t) = \delta h_2(t), \;\; \delta \ne 1
                        \end{split}
                    \end{equation}
                
                The test statistic for the log-rank test is calculated as the sum of the differences between the observed and expected number of events squared, divided by the expected number of events, summed over all observed time points. The p-value obtained from the log-rank test indicates the significance of the difference in survival distributions between the two groups. The test statistic is chi-squared under the null hypothesis \cite{Lifelines}.  

                    \begin{equation}
                        \chi^2 = \sum_{i=1}^{N} \frac{(O_i - E_i)^2}{E_i}
                    \end{equation}
                    where $O_{i}$ is the observed number of events at time point $i$ in the sample, $E_{i}$ is the expected number of events at time point $i$ under the null hypothesis, and $N$ is the total number of observed time points.

                \item \textbf{Huber Loss}: For TLS ratio prediction, we used Huber Loss, a loss function commonly used in regression tasks, known for combining the advantages of both the Mean Absolute Error (MAE) and the Mean Squared Error (MSE). It behaves differently based on the magnitude of the error; it is quadratic for small errors and linear for large errors. This characteristic makes it less sensitive to outliers than MSE and more sensitive to small errors than MAE. Huber loss function is defined as follows:                

                \begin{equation}
                    L_{n} =
                    \begin{cases}
                    0.5 (y_{n} - \hat{y_{n}})^2, & \text{if } |y_{n} - \hat{y_{n}}| \leq \delta, \\
                    \delta * \left( |y_{n} - \hat{y_{n}}| - 0.5 * \delta \right), & \text{otherwise}.
                    \end{cases}
                    \end{equation}
                
                where \(( y - \hat{y} ) \) represents the residual, which is the difference between the actual value and the predicted value, and \( \delta \) is a positive threshold parameter that determines the point at which the loss function transitions from quadratic to linear behavior \cite{Huber}.
                
                \item \textbf{Wilcoxon Signed-Rank Test}: To assess the agreement between the manually annotated TLS ratio and the model's predictions, we used the Wilcoxon Signed-Rank test. This non-parametric statistical method evaluates whether there is a significant difference between the paired values, taking into account both the magnitude and direction of the differences. The null hypothesis assumes that the two distributions are statistically similar. A two-sided p-value of less than 0.05 was considered evidence of a significant difference between the two sets of ratios.
            \end{enumerate}
        
        \begin{table}[ht]
        \centering
        \caption{Hyperparameters search for training.}
        \begin{tabular}{>{\raggedright}p{3.2cm}>{\raggedright\arraybackslash}p{5.2cm}}
        \toprule
            \textbf{Hyperparams} & \textbf{Training (range)} \\ \midrule
            Learning Rate   & [1e-6, 1e-1]          \\ 
            Weight Decay    & [1e-6, 1e-1]          \\
            Dropout         & [0.1, 0.65]           \\ 
            Batch Size      & [64, 128, 256, 512]   \\
            Epochs          & [50, 100]             \\
            Hidden Layers   & [1, 2, 3, 4, 5, 6, 7, 8, 9]    \\ 
            Hidden Neurons  & [2048, 1024, 512, 256, 128, 48, 32]   \\ 
            Optimizer       & [adam, sgd, rmsprop, adamw]   \\
            Learning Rate Policy & [linear, exp, step, plateau, cosine] \\ \hline
        \end{tabular}
        \label{table:hyperparams}
        \end{table}

        \begin{table}[ht]
            \centering
            \caption{Frameworks and packages used in our codebase.}\label{tabS4}
                \begin{tabular}{l l c}
                \toprule
                      & {{\textbf{Package name}}} &   {{\textbf{Version}}} \\
                \midrule
                \textbf{Operating systems}        &   Ubuntu         & 20.04.4 \\            
                \hline
                \textbf{Programming languages}    &   Python         & 3.10.13  \\
                \hline
                \textbf{Deep learning framework}  &   Pytorch        & 2.2.0  \\
                                                  &   torchvision    & 0.17.0  \\
                \hline
                                                  &   feature-engine & 1.6.2   \\
                                                  &   imbalanced-learn   & 0.12.0   \\
                \textbf{Miscellaneous}            &   scipy          & 1.12.0    \\
                                                  &   scikit-learn   &   1.4.0   \\
                                                  &   numpy          & 1.26.3   \\
                                                  &   PyYaml         & 6.0.1    \\
                                                  &   jupyter        & 1.0.0    \\
                                                  &   pandas         &   2.2.0   \\
                                                  &   pickle5        &   0.0.11  \\
                                                  &   protobuf       &   4.25.2  \\
                                                  &   wandb          &   0.16.3  \\                                                
                \bottomrule
                \end{tabular} \label{table:frames}
        \end{table}

        \subsubsection{Hyperparameters Search}
        Hyperparameters are non-learnable parameters of a deep learning model and are crucial as they govern the learning process and model architecture. Hyperparameter tuning involves selecting the optimal combination of parameters that results in the best model performance. Common hyperparameters include learning rate and policy, batch size, number of epochs, weight decay, dropout type and probability, and architecture specifics such as the number of hidden layers and neurons in each layer. Methods for hyperparameter search range from grid search, where all possible combinations of parameters are evaluated; to random search, which randomly samples parameter combinations within predefined bounds. More sophisticated techniques like Bayesian optimization or using automated machine learning (AutoML) tools can dynamically adjust parameters based on previous results to find the best solutions more efficiently. We employed weights and biases \cite{wandb} utility to carry out random and Bayesian methods of hyperparameters search. The list of hyperparameters we searched for training is given in Table \ref{table:hyperparams}. For model training, we conducted around 400 simulations to find the current hyperparameters. To further verify the performance of our model, we evaluated the model with the off-the-shelf datasets CPTAC-LSCC \cite{CPTAC-LSCC} and Moffitt's LSCC \cite{stewart2019proteogenomic}. The plot for these simulations is given in Figure \ref{suplefig:1}.

        \subsubsection{Frameworks, Compute resources, and wall-clock times}
        We trained SeNMo model using the Moffitt Cancer Center's HPC machine using one Tesla V100 32GB GPU running Ubuntu 22.04.4 and CUDA 12.2. The entire code was developed in Python and PyTorch frameworks. The software frameworks and corresponding packages used in our codebase are given in Table \ref{table:frames}. Training time for our current 83.33 Million parameter SeNMo encoder is approximately 11 hours. We conducted the hyperparameters search of the pan-cancer model for approximately 20 days using multiple GPUs in parallel. Finetuning the trained model on a given data having around 150 patients approximately takes 15 minutes. 

        \begin{figure}[ht!]%
    \centering
    \includegraphics[width=1.0\textwidth]{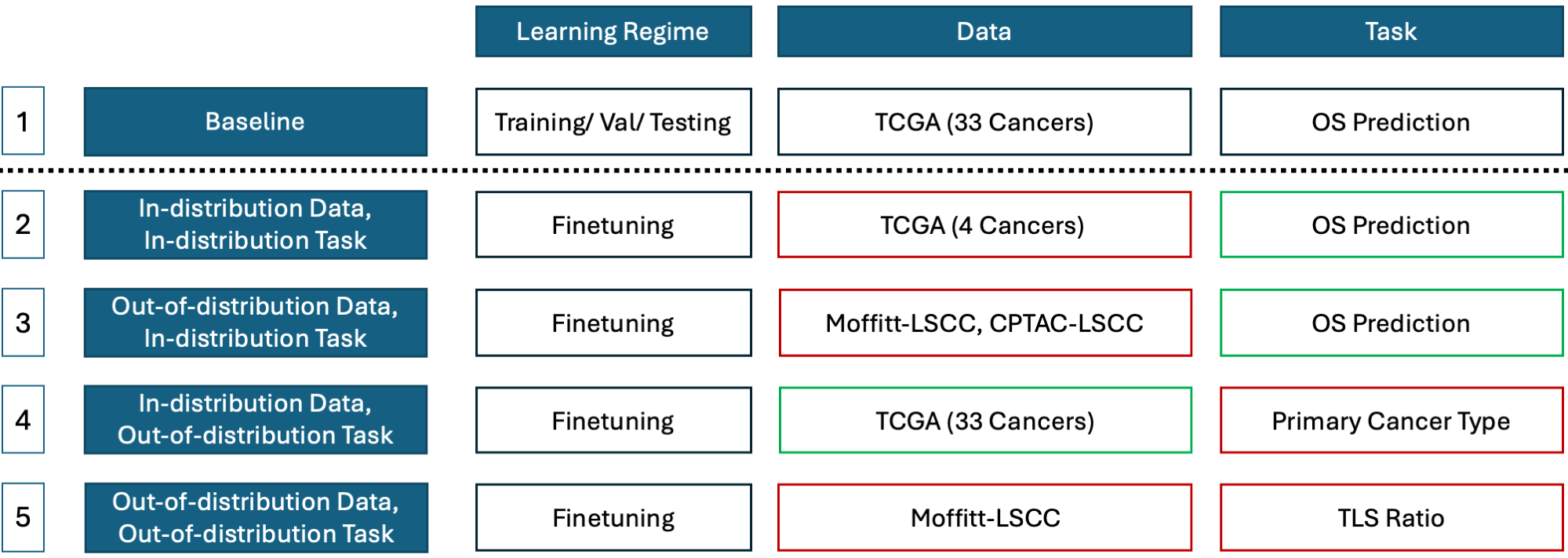}
    \caption{Study design and simulations structure for SeNMo across different learning regimes, datasets, and tasks. The baseline model (row 1) is first trained on TCGA data with 33 cancer types for overall survival (OS) prediction. Rows 2–5 represent variations: red-bordered boxes indicate a change from the baseline (e.g., out-of-distribution task and/or data), while green-bordered boxes align with the baseline. Simulations include OS prediction on both seen and unseen data (rows 2 and 3) and new tasks such as primary cancer type classification and TLS ratio prediction on seen and unseen datasets (rows 4 and 5).}\label{fig:work-layout}
\end{figure}

\subsection{Study Design}
An overview of the various simulations conducted to evaluate the capabilities of the SeNMo model across different learning regimes, tasks, and datasets is shown in Figure \ref{fig:work-layout}. The study design included multiple learning regimes, each designed to assess the model’s adaptability, generalizability, and robustness. The baseline model was initially trained on TCGA dataset comprising 33 different cancer types for OS prediction. The subsequent learning regimes explored different data variations and tasks, which we call out-of-distribution simulations because the model had not encountered such data/task in baseline learning. These scenarios included OS prediction on both seen and unseen datasets, as well as tasks such as primary cancer type classification on seen data and TLS ratio prediction on unseen data.

\begin{figure}[ht!]%
    \centering
    \includegraphics[width=1.0\textwidth]{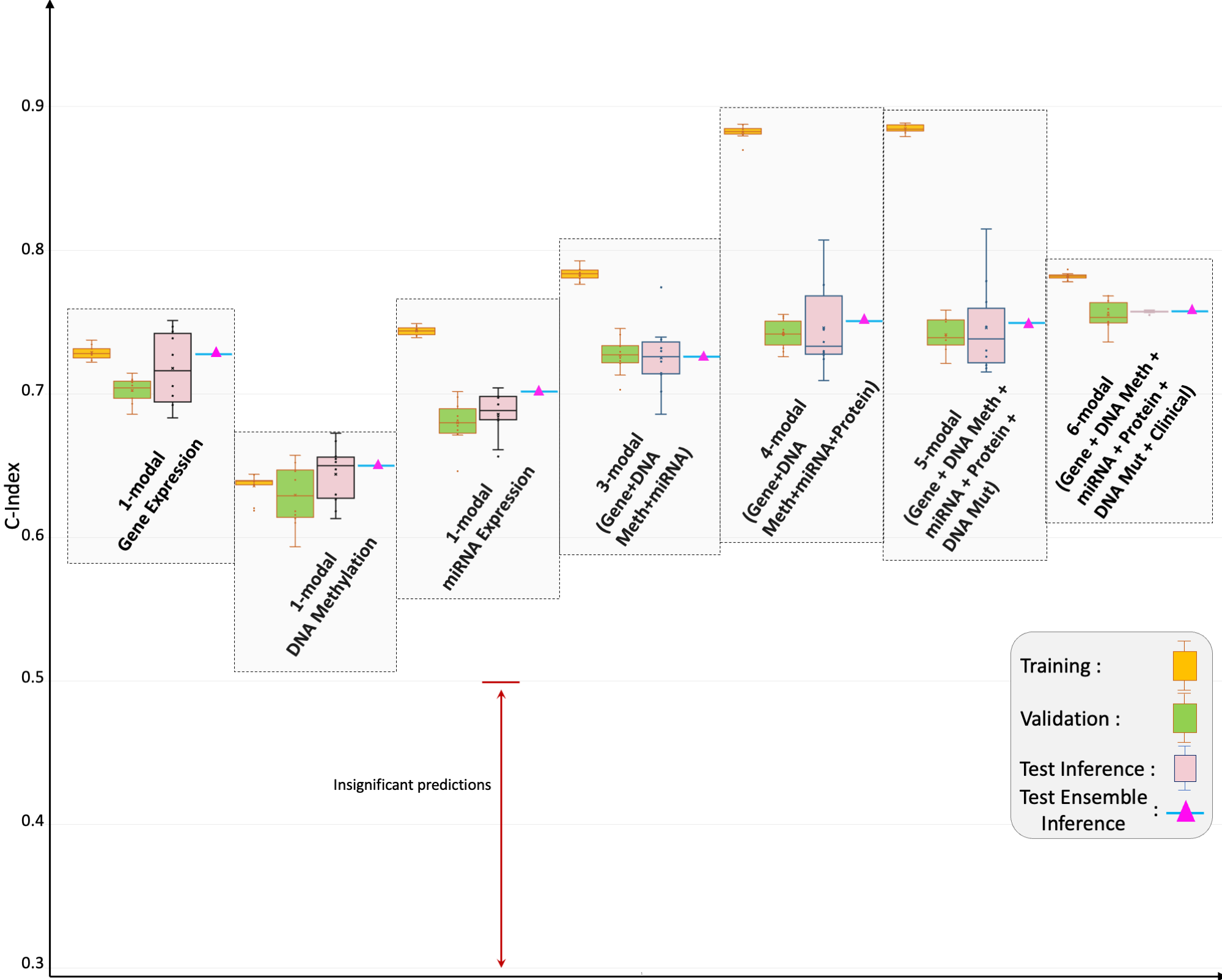}
    \caption{Pan-cancer C-Index results for OS prediction. The SeNMo model was trained and evaluated using different combinations of data modalities. Training and validation were carried out on the $80\%$ of the total data, whereas inference was done on the $20\%$ held-out test set. As the number of modalities increased in the pan-cancer data, the model's performance improved, as depicted by the upward trend of C-Index. All the results shown here are statistically significant, i.e., $p<0.05$.}\label{fig:panCindx}
\end{figure}

\section{Results}
\subsection{Pan-Cancer Multimodal Analysis for Predicting Overall Survival}

Figure \ref{suplefig:1} shows the visualization of the parallel sweeps across all hyperparameters, resulting in training around $400$ unique models. The optimal model had a learning rate of $0.00058$, a weight decay of $0.00598$, $0.1058$ dropout, $256$ batch size, $100$ epochs, and seven hidden layers with neurons in these layers as $[1024, 512, 256, 128, 48, 48, 48]$. The trained model contained $83.33$ million trainable parameters. Checkpoints were saved for this model for each of the $10$ folds. The model's training resulted in the average training C-Index of $0.78$ and average validation C-Index of $0.76$ across the $10$ folds. The inference on the test set showed the C-Index of $0.757$, the average of the C-Indices from the $10$ checkpoints. To further validate our findings, we created an ensemble of the $10$ checkpoints by averaging the prediction vectors from all the models and then evaluating the final averaged prediction vector for C-Index. For the pan-cancer, multi-omics SeNMo model, an ensemble C-Index of $0.758$ was achieved on the held-out test set. The significance level in all these analyses is $95\%$, i.e., $p<0.05$, indicating statistically significant values. These results are depicted in the Figure \ref{fig:panCindx}. 

As depicted in Figure \ref{fig:panCindx}, the SeNMo model trained on the pan-cancer 1-modal (Gene expression) cohort showed a C-Index for training, validation, testing, and ensemble inference as $0.729, 0.702,	0.718,$	and $0.728$, respectively. For the pan-cancer 1-modal (DNA methylation) cohort, the model's training, validation, testing, and ensemble inference C-indices are $0.636,	0.629,	0.644,$ and $0.65$, respectively. For the pan-cancer 1-modal (miRNA expression) cohort, the model's training, validation, testing, and ensemble inference C-indices are $0.744, 0.68, 0.686,$ and $0.702$, respectively. We did not analyze the model individually on the rest of the three modalities because clinical and protein expression features are too small for an $83$ million-parameter model, whereas the DNA mutation data comprised the binarized features of mutations. Evaluating the model on the 3-modal cohort showed the training, validation, testing, and ensemble inference C-indices of $0.783, 0.727, 0.725,$ and $0.726$, respectively. Further adding the protein expression to the 3-modal data, we trained and evaluated the model on the 4-modal cohort and got the C-Indices of $0.88, 0.742, 0.746,$ and $0.751$ for training, validation, testing, and ensemble inference, respectively. Lastly, the model's performance on the 5-modal cohort showed the training, validation, testing, and ensemble inference C-indices of $0.885, 0.741, 0.746,$ and $0.749$, respectively. Next, we analyze how the model trained on pan-cancer, 6-modal data fared on individual cancer patients' data.

\begin{figure}[ht!]%
    \centering
    \includegraphics[width=1.0\textwidth]{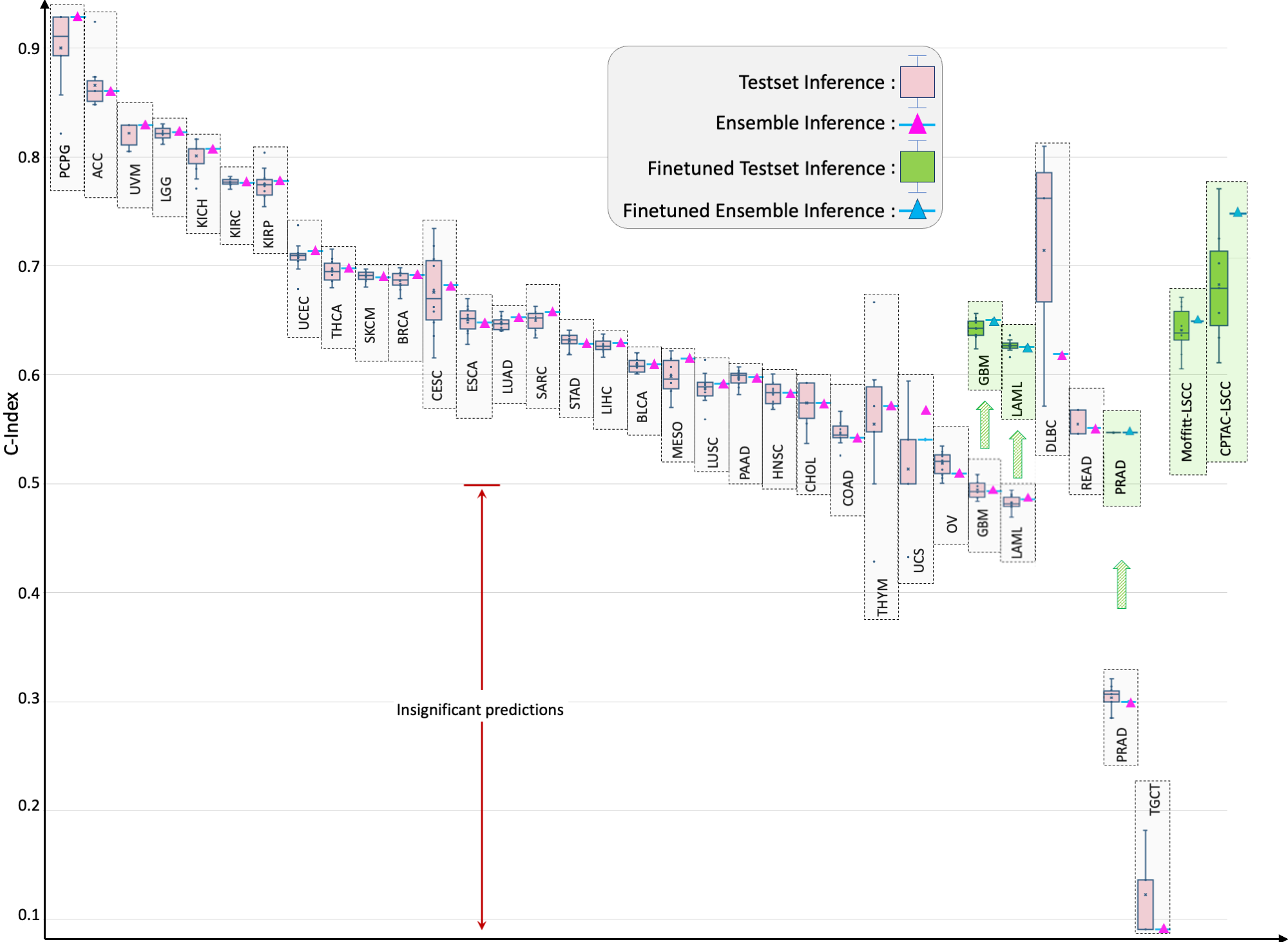}
    \caption{Cancer-specific C-index results for overall survival (OS) prediction across multiple cancer types. Pink box plots represent predictions on the held-out test set, while green box plots show predictions from fine-tuned models for cancer types with initially insignificant results or for unseen data, such as Moffitt LSCC and CPTAC-LSCC. Triangular markers indicate ensemble predictions, with blue triangles representing fine-tuned ensemble results. Models with insignificant predictions fall below the 0.5 threshold, marked by a red arrow. Although trained on pan-cancer cohort, SeNMo effectively captures survival times across individual cancers, with fine-tuning improving performance for cases with initially low predictive significance.}\label{fig:cancersCindx}
\end{figure}

\subsection{Individual Cancer Multimodal Analysis for Predicting Overall Survival}
We evaluated the model trained on the 6-modal pan-cancer cohort on the held-out individual cancer data from an individual cancer-wise perspective. The number of patients in these cancer cohorts was a randomly selected subset of the cases shown in Figure \ref{fig:cases} and Tables \ref{Tab:features}, \ref{Tab:clinchars}, which accounts for the $20\%$ of the total samples. The trained model was evaluated on each of the 33 individual cancer data using simple inference and the ensemble of the 10-fold checkpoints. Figure \ref{fig:cancersCindx} shows the evaluation performance of the model on 33 cancer types. The model showed the best predictive performance on TCGA-PCPG data with an average C-Index on the test set of $0.9$ and ensemble inference of $0.929$. SeNMo's performance on the other cancer types in format $\{$\emph{Test Inference}, \emph{Ensemble Inference}$\}$\ is shown in Table \ref{tab:cinf}, where 29 cancer types have significant C-Indices. We noticed that the results for TCGA-GBM, TCGA-LAML, TCGA-PRAD, and TCGA-TGCT were not statistically significant, i.e., $p>0.05$. So, we fine-tuned the model for these datasets by reducing the learning rate, increasing the weight decay and dropout, and letting the model fine-tune for 10 epochs. Resultantly, the model's performance increased for TCGA-GBM$=\{0.642, 0.650\}$, TCGA-LAML$=\{0.627, 0.626\}$, and TCGA-PRAD$=\{0.541, 0.542\}$. These improvements are depicted with the green arrows and green boxes in Figure \ref{fig:cancersCindx}. However, the model failed to converge for TCGA-TGCT data and consistently gave predictions that were not significant, $p>0.05$.

    \begin{table}[ht]
    \centering
    \caption{C-Index for Test and Ensemble Inference across Cancer Types.}
    \small
    \begin{tabular}{p{2.5cm} p{2.5cm} | p{2.5cm} p{2.5cm}}
    \toprule
    \textbf{Cancer Type} & \makecell{\textbf{C-Index} \\ \{Test, Ensemble\}} & \textbf{Cancer Type} & \makecell{\textbf{C-Index} \\ \{Test, Ensemble\}} \\
    \midrule
    TCGA-PCPG & \{0.900, 0.929\} & TCGA-BLCA & \{0.609, 0.609\}  \\
    TCGA-ACC & \{0.866, 0.861\} & TCGA-MESO & \{0.599, 0.615\} \\
    TCGA-UVM & \{0.822, 0.829\} & TCGA-LUSC & \{0.588, 0.592\}\\
    TCGA-LGG & \{0.821, 0.823\} & TCGA-PAAD & \{0.597, 0.598\}\\
    TCGA-KICH & \{0.801, 0.807\} & TCGA-HNSC & \{0.583, 0.583\}  \\
    TCGA-KIRC & \{0.777, 0.776\} & TCGA-CHOL & \{0.574, 0.574\} \\
    TCGA-KIRP & \{0.775, 0.778\} & TCGA-COAD & \{0.546, 0.542\}\\
    TCGA-UCEC & \{0.708, 0.713\} & TCGA-THYM & \{0.555, 0.571\} \\
    TCGA-THCA & \{0.696, 0.698\} & TCGA-UCS & \{0.514, 0.541\} \\
    TCGA-SKCM & \{0.691, 0.689\} & TCGA-OV & \{0.518, 0.509\}\\
    TCGA-BRCA & \{0.687, 0.692\} & TCGA-GBM & \{0.495, 0.493\} \\
    TCGA-CESC & \{0.676, 0.682\} & TCGA-LAML & \{0.482, 0.485\} \\
    TCGA-ESCA & \{0.650, 0.648\} & TCGA-DLBC & \{0.714, 0.619\} \\
    TCGA-LUAD & \{0.647, 0.653\} & TCGA-READ & \{0.550, 0.551\} \\
    TCGA-SARC & \{0.650, 0.658\} & TCGA-PRAD & \{0.304, 0.300\} \\
    TCGA-STAD & \{0.631, 0.628\} & TCGA-TGCT & \{0.123, 0.091\} \\
    TCGA-LIHC & \{0.627, 0.629\} & & \\
    \bottomrule
    \end{tabular}
    \label{tab:cinf}
\end{table}

\subsection{Out-of-distribution Evaluation and Fine-tuning}
Evaluating the model without fine-tuning showed the $\{$\emph{Test Inference}, \emph{Ensemble Inference}$\}$\ of CPTAC-LSCC$=\{0.48, 0.50\}$, and Moffit-LSCC$=\{0.581, 0.59\}$. Fine-tuning the model for 10 epochs, with reduced learning rate, and increased weight decay and dropout resulted in the improvement of C-Indices as CPTAC-LSCC$=\{0.677, 0.73\}$, and Moffit-LSCC$=\{0.647, 0.656\}$. These fine-tuning results are depicted in Figure \ref{fig:cancersCindx} as the green box plots.

\begin{figure}[ht!]%
    \centering
    \includegraphics[width=1.0\textwidth]{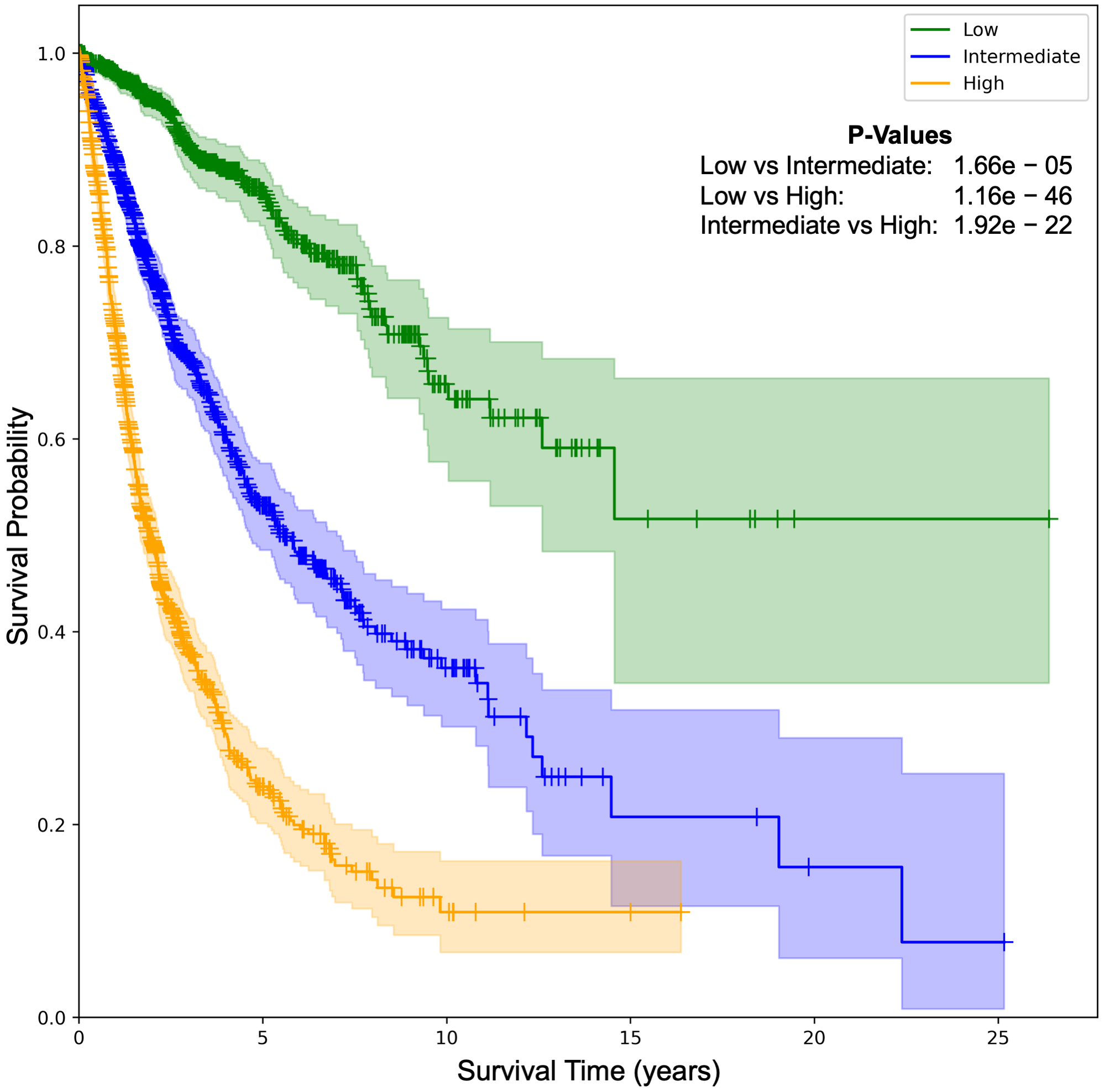}
    \caption{Kaplan-Meier (KM) comparative analysis of using SeNMo in stratifying patient outcomes in low/ intermediate/ high risk, defined by the 33-66-100 percentile of hazard predictions. Hazard predictions from SeNMo show clear distinctions between stratified groups. The p-values from logrank test for Low vs. Intermediate: $1.66e-05$, Low vs. High: $1.156e-46$, and Intermediate vs. High: $1.92e-22$. The shaded areas around each curve depicts the 95\% confidence intervals.}\label{fig:KM}
\end{figure}

\subsection{Patient Stratification}
We further investigated the SeNMo's ability to stratify the patients based on low, intermediate, and high risk conditions. We generate Kaplan-Meier (KM) curves of our model on the pan-cancer, multi-omics held-out test set, as shown in Figure \ref{fig:KM}. We select the low/ intermediate/ high risk stratification distribution as the 33-66-100 percentile of hazard predictions \cite{chen2020pathomic, li2023survival}. The hazard scores predicted by SeNMo are used to evaluate the model's stratification ability. The KM comparative analysis shows that SeNMo distinguished the patients across the three groups. The low-risk group (green) exhibited the highest survival probability, maintaining close to 100\% survival up to approximately 5 years, and gradually declining to about 60\% by the 25-year mark. The intermediate-risk group (blue) showed a significantly lower survival probability, starting to diverge from the low-risk group early on and reaching around 40\% by the 15-year mark of the study period. The high-risk group (orange) displayed the most pronounced decline in survival probability, with a steep drop to approximately 20\% survival within the first 10 years, and further reducing to below 10\% after 10 years. The logrank test to evaluate the significance of this stratification shows that the p-value of low vs. intermediate curves is $1.66e-05$, low vs. high is $1.156e-46$, and intermediate vs. high is $1.92e-22$, showing significant results, i.e., $p<0.05$. The 95\% confidence intervals around each curve show the reliability of these estimates.


\subsection{Primary Cancer Type Prediction}
\begin{figure}[ht!]%
    \centering
    \includegraphics[width=1.0\textwidth]{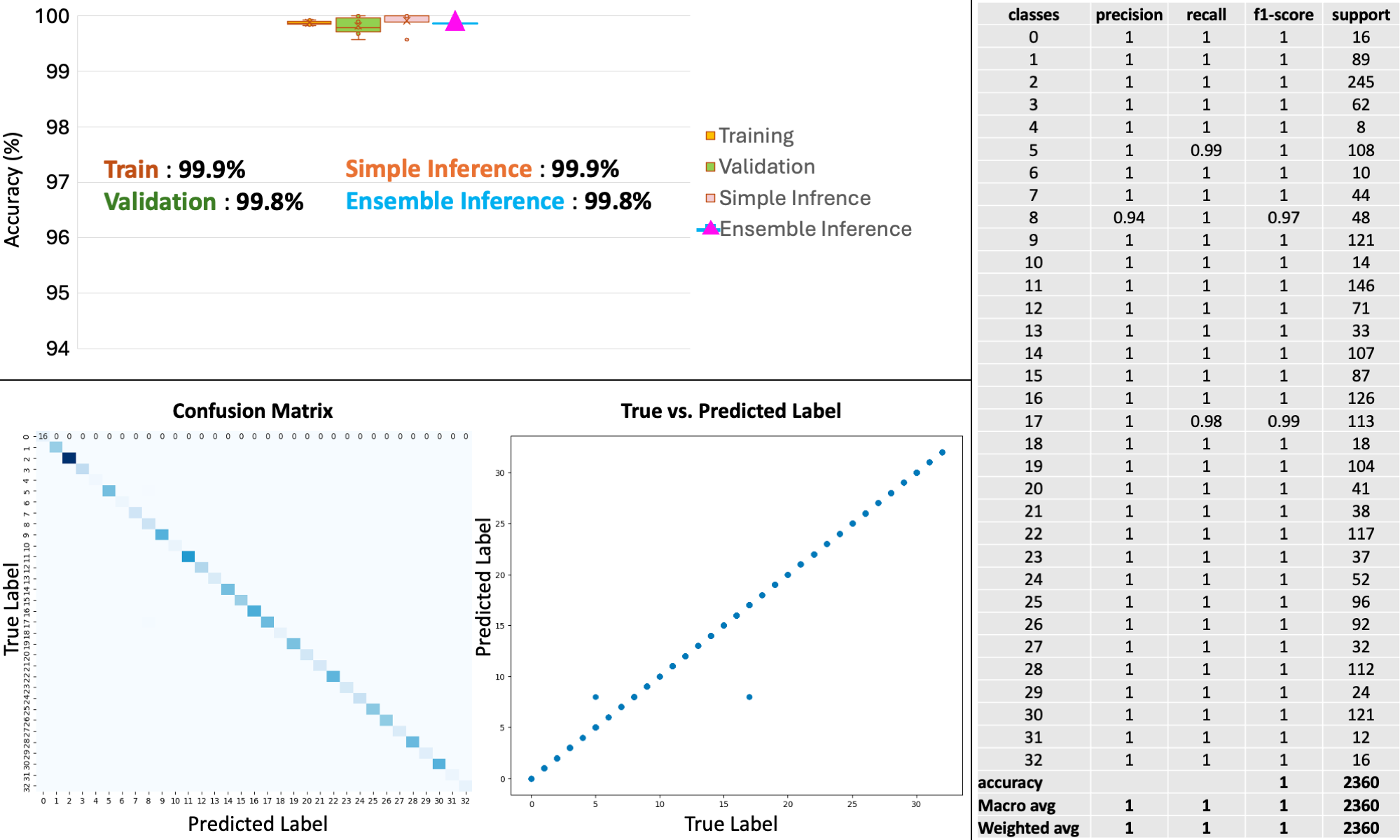}
    \caption{Pan-cancer primary cancer type prediction results. The model's accuracy across training, validation, and inference stages is near-perfect (top left panel). Confusion matrix (bottom left) shows minimal misclassifications, while the scatter plot (bottom middle) shows the alignment of predicted versus true labels. The classification report (right panel) shows high precision, recall, and f1-scores in the 33 cancers type-identification.}\label{fig:classification}
\end{figure}

To test the generalizability of SeNMo across different tasks, we carried out the prediction of primary cancer type from pan-cancer, multi-omics data. We set the problem as a classification problem, where the multi-omics data is used to predict the type of cancer for the given patient data among the 33 classes. It is imperative to mention here that the four clinical features in the initial data contained the cancer stage, as shown in Figure \ref{fig:preprocess} and Table \ref{Tab:clinchars}. When considering a cancer type classification problem, the stage adds a bias in the data because of the staging distribution among different cancers. Therefore, for the cancer classification simulations, we excluded the ``stage'' feature in the clinical data. As shown in Figure \ref{fig:classification}, the model achieves near-perfect accuracy levels, with 99.9\% average accuracy in training, 99.8\% in validation, and consistent performance in both simple and ensemble inference approaches. The confusion matrix depicts a clear concentration of values along the diagonal, indicating a high rate of correct predictions across all cancer types. The scatter plot shows an alignment of predicted labels with true labels along the diagonal line, highlighting the model’s robust predictive accuracy. The classification report across various cancer types reveals that the model consistently maintains high precision, recall, and F1-scores, approaching a value of 1 for almost all categories. The robust predictive power of our model emphasizes the fact that each cancer has a unique molecular landscape, highlighted through differences in gene, protein, and miRNA expression, DNA methylation, and types of somatic mutations seen in our data.

\begin{figure}[ht!]%
    \centering
    \includegraphics[width=0.9\textwidth]{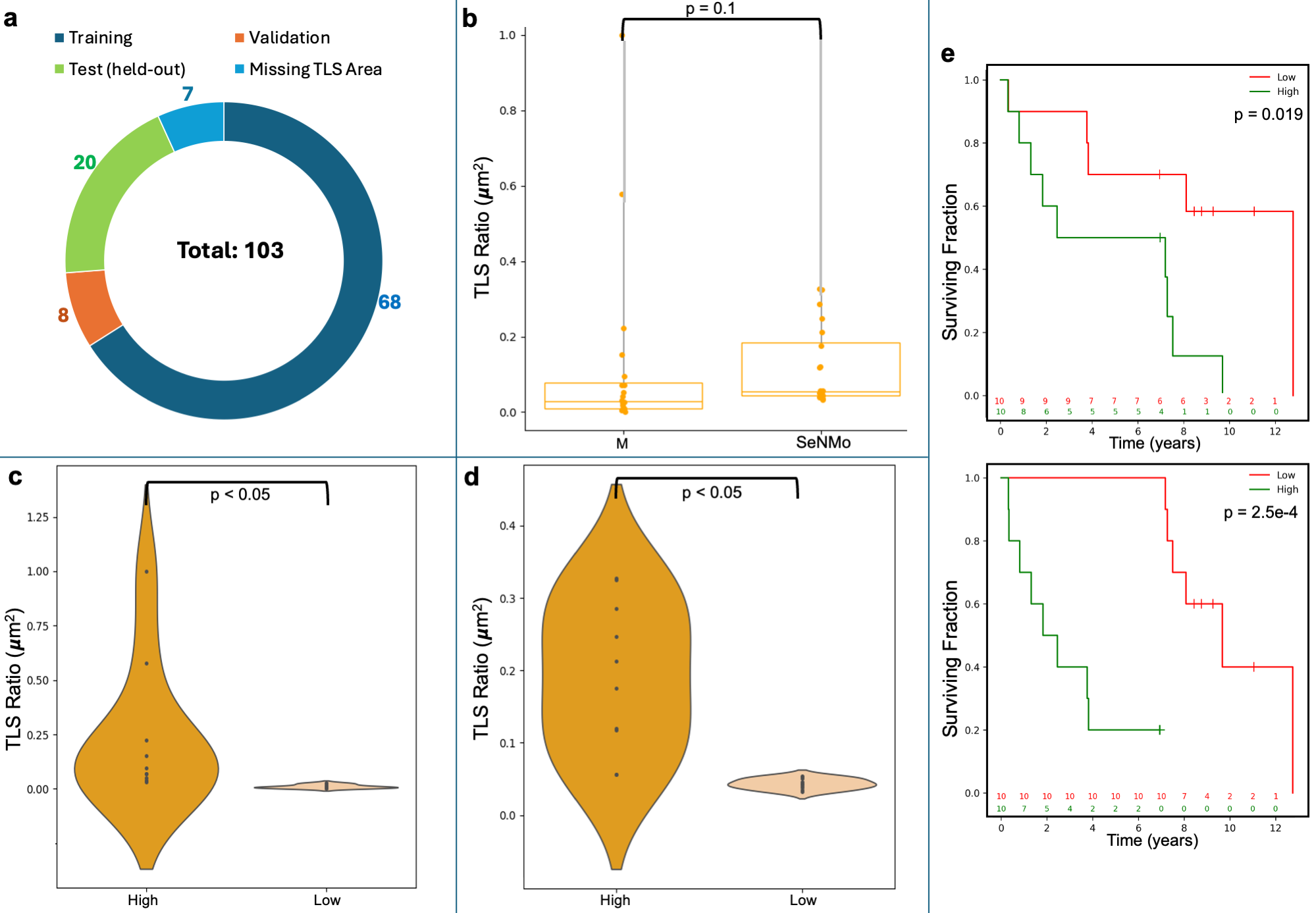}
    \caption{Tertiary Lymph Structures (TLS) Ratio Predictions by SeNMo. (\textbf{a}) Dataset distribution, dividing 103 samples into training, validation, test, and excluding those missing TLS data. (\textbf{b}) Box plots compare TLS ratios between manual annotations (M) and SeNMo predictions, showing no significant difference ($p=0.1$). (\textbf{c}) Violin plot of TLS Ratio annotations for high vs. low groups thresholded by the median value, with a significant difference ($p<0.05$). (\textbf{d}) Violin plot of SeNMo's TLS Ratio predictions, also showing significant separation between high and low groups ($p<0.05$). (\textbf{e}) Kaplan-Meier survival curves with significant survival differences, comparing high vs. low TLS ratios for both annotations (top, $p=0.019$) and SeNMo predictions (bottom, $p=2.5e-4$).}\label{fig:TLS}
\end{figure}

\subsection{Tertiary Lymph Structures (TLS) Ratio}
To further evaluate SeNMo's generalizability on previously unseen data and across different tasks, we fine-tuned the model to predict the TLS ratio on a cohort of lung squamous cell carcinoma data collected at Moffitt Cancer Center. This task was formulated as a regression problem. As shown in Figure \ref{fig:TLS}, the TLS ratio predictions generated by SeNMo demonstrated strong performance on the held-out test set. Specifically, the comparison between manual TLS ratio annotations and SeNMo-predicted ratios revealed no significant difference ($p=0.1$), indicating a high level of concordance between manual assessments and model predictions (Figure \ref{fig:TLS}b). Further analysis using violin plots compared the distribution of TLS ratios for manually annotated high vs. low groups with those predicted by SeNMo. Both manual and predicted TLS ratios showed significant separation between high and low groups ($p<0.05$), highlighting the model's ability to accurately distinguish between different levels of TLS (Figures \ref{fig:TLS}c and \ref{fig:TLS}d). Moreover, KM survival analysis was performed to assess the prognostic value of TLS ratios. Survival curves revealed significant differences in survival outcomes between patients with high and low TLS ratios, both for manually annotated data ($p=0.019$) and SeNMo-predicted data ($p=2.5e-4$) (Figure \ref{fig:TLS}e).

\section{Discussion}
\begin{figure}[H]%
        \centering
        \includegraphics[width=1.0\textwidth]{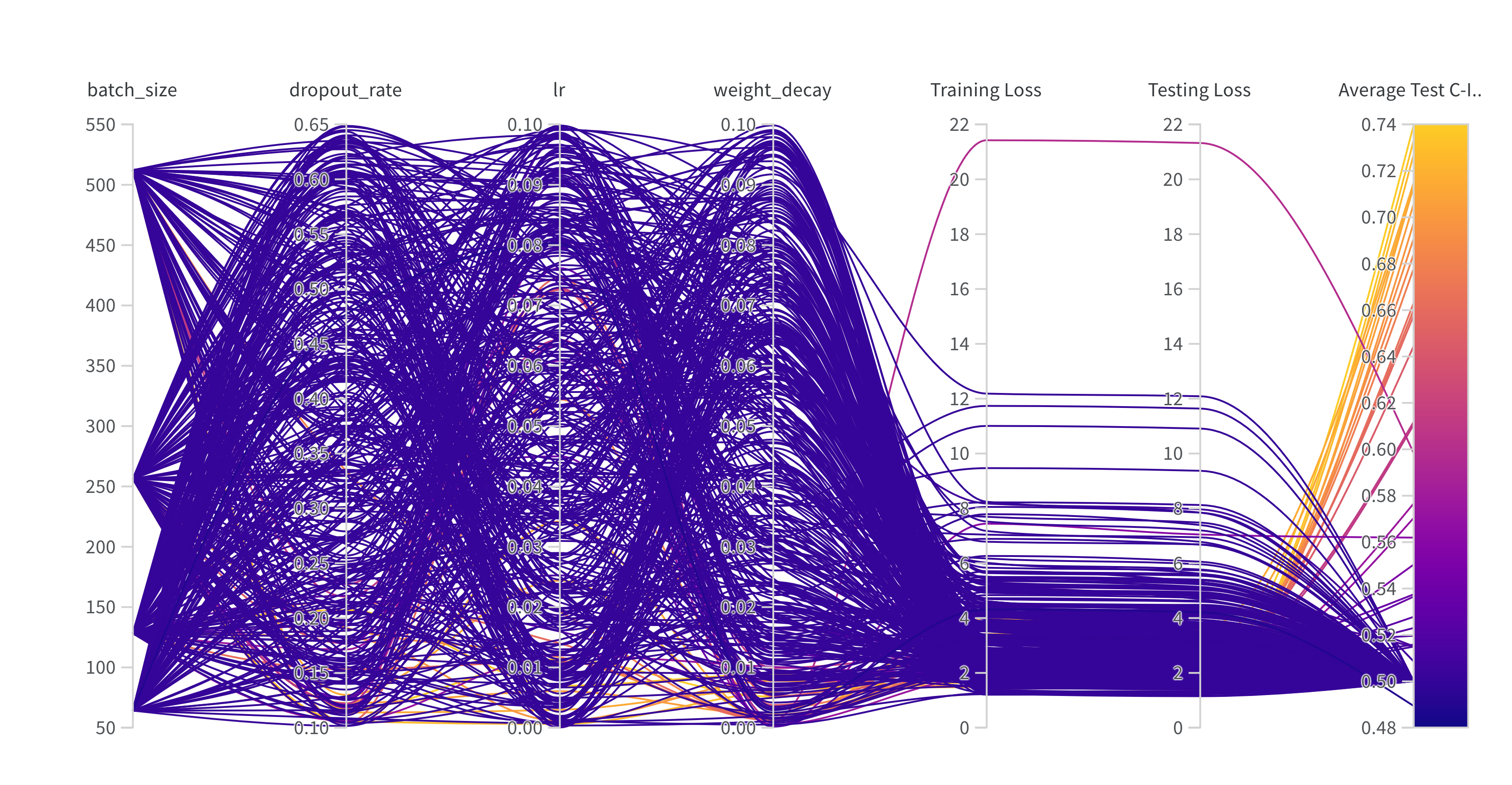}
        \caption{Hyperparameters search for training the SeNMo model on Pan-cancer multiomics data. The goal here was to maximize the validation C-Index.}\label{suplefig:1}
    \end{figure}

We analyzed pan-cancer dataset of 33 cancer types comprising five molecular data modalities (with varying amount of features) and four clinical data features using our SeNMo encoder-based framework. Public databases such as CPTAC and TCGA contain common identifiers within their data that connect data from the same patient. Therefore, molecular data, such as gene expression, miRNA expression, DNA methylation, somatic mutations, and protein expression can be consolidated to represent a singular patient. However, such high-dimensional data has intra- and inter-dataset correlations, heterogeneous measurement scales, missing values, technical variations, and other forms of noise \cite{zhao2024tutorial}. This necessitates the need for a variety of preprocessing techniques such as the removal of low variance features and the imputing of missing features among others prior to training. Training such a large dataset having high-dimensional heterogeneous data required proper computational resources and a precise pipeline for training, testing, and validation.  After extensive training-evaluation runs, we found, through optimal parameters searching, a model that performs very well across the different data types and tasks (refer to Figures \ref{fig:panCindx} and \ref{suplefig:1}).  The model has been shown to outperform the existing works in OS prediction when considering the six data modalities included in our data \cite{nikolaou2024quantifying}. Moreover, we observed that adding more data and types of modalities increased the model's performance. 

The model's performance was evaluated on individual cancers at test-time through simple inference and ensembling methods. We observed that the model's predictive power improved when an ensemble of the checkpoints was employed, (refer to Figure \ref{fig:cancersCindx}). However, for four cancer types, TCGA-GBM, TCGA-LAML, TCGA-PRAD, and TCGA-TGCT, the model did not show significant predictive power. During the investigation, we observed that these datasets had non-admissible pairs in some of the data folds, i.e., all samples had censor value $\delta=0$ in Equation \ref{eq:9}. In the case of TCGA-PRAD and TCGA-TGCT, the number of samples having $\delta=1$ in the training/validation cohort was $12$ and $3$, respectively. To address the lack of predictive power, we fine-tuned the model for these datasets by using the stratified k-folds to offset the class-representation problem in the data folds. After searching for the optimal hyperparameters for fine-tuning, the model's performance became significant ($p<0.05$) for three out of four datasets, (refer to green box plots in Figure \ref{fig:cancersCindx}).

It is imperative to mention here that MLPs-based networks are very sensitive to catastrophic forgetting when presented with out-of-distribution data or when subjected to a different task \cite{liu2024kan}. We fine-tuned the SeNMo encoder for one public data (CPTAC-LSCC) and one internal data (Moffitt's LSCC) \cite{CPTAC-LSCC, stewart2019proteogenomic}. In our simulations to fine-tune the model, we encountered the catastrophic forgetting phenomenon in SeNMo, where the model would fail to converge on both new datasets. This was more pronounced when a certain number of hidden layers were frozen, and the rest were trained with lower learning rates. We resorted to the option of unfreezing all the layers of the encoder and fine-tuning the model with a very small learning rate ($4e-5$), high weight decay and dropout ($0.35$), and just $10$ epochs. This method worked and the model showed significant performance on the out-of-distribution datasets.

Risk stratification of patients allows clinicians and researchers to identify patients who might need more intensive care or monitoring and those who may have a better prognosis, facilitating more personalized treatment approaches. The KM survival curves depicted in Figure \ref{fig:KM} demonstrate a clear stratification of survival probabilities among three risk-defined patient groups. These results underscore the effectiveness of the risk stratification model in predicting long-term outcomes and highlight the critical need for targeted therapeutic strategies based on individual risk assessments. This stratification allows for more personalized patient management and could potentially guide clinical decision-making toward improving OS rates across diverse patient populations.

Cancer type classification is routinely studied for early detection and localization of tissue of origin \cite{gore2022cancernet}. The classification results in Figure \ref{fig:classification} illustrate the superior generalizability of the model's predictive power to classify primary cancer types through the SeNMo encoder, despite it being primarily trained for predicting OS. Additionally, the detailed classification report across various cancer types reveals that the model consistently maintains high precision, recall, and F1-scores for almost all cancer types. Such metrics not only confirm the model's effectiveness in accurately identifying the correct cancer class but also its reliability in replicating these results across different samples. This level of performance suggests the capability of the model to successfully learn high level representations from heterogenous, high-dimension, mutlivariate data stemming from complex molecular modalities such as gene expression, miRNA expression, somatic mutations, DNA methylation, and protein expression. 

As shown in Figure \ref{fig:TLS}, SeNMo's ability to predict TLS ratios was evaluated on an unseen cohort of lung squamous cell carcinoma data from Moffitt Cancer Center. The comparison between manual TLS ratio annotations and SeNMo-predicted values showed no significant difference ($p=0.1$), indicating a high level of concordance between human annotations and model predictions. Violin plots depicting high vs. low TLS ratio groups—both for manual and SeNMo predictions—revealed significant separation ($p<0.05$), demonstrating the model's robustness in distinguishing between biologically distinct TLS levels. Furthermore, KM survival curves for high vs. low TLS ratio groups revealed significant differences in survival outcomes, with stronger statistical significance observed for SeNMo-predicted data ($p=2.5e-4$) compared to manual annotations ($p=0.019$). These results underscore the potential of SeNMo to not only replicate expert-driven TLS annotations but also provide a consistent and potentially superior prognostic assessment. Overall, the results indicate that SeNMo can successfully generalize to new tasks and datasets, accurately predicting TLS ratios and offering valuable prognostic insights that could improve clinical decision-making.

We made the entire codebase of SeNMo publicly available on GitHub (\url{https://github.com/lab-rasool/SeNMo}). We have made the latent representations of patient data generated from SeNMo available to the research community through our HoneyBee system \cite{tripathi2024honeybee}. HoneyBee stores these representations, also known as patient embeddings, in a structured format using Hugging Face datasets, effectively creating a vector database. HoneyBee has demonstrated the effectiveness of using patient embeddings, offering a significant advantage over the traditional approach of using raw data and extensive pre-processing \cite{tripathi2024honeybee}.

\section{Conclusion}
In this study, we introduced SeNMo, a foundational deep learning model specifically designed for multi-omics data analysis across 33 different cancer sites. By leveraging high-dimensional multi-omics datasets from the NCI Genomics Data Commons, SeNMo demonstrated robust performance in predicting overall survival on both training and held-out test sets. The model’s adaptability and efficiency were further validated through its high accuracy in classifying primary cancer types and predicting TLS ratios, showcasing its ability to generalize effectively across different tasks. As a foundational model, SeNMo represents a resilient and scalable solution that advances the integration and analysis of complex molecular data, providing a comprehensive understanding of cancer biology. Our approach underscores the potential of self-normalizing networks in oncology, emphasizing the importance of comprehensive data preprocessing and optimal parameter tuning. By making SeNMo and its derived patient embeddings publicly available, we aim to facilitate further research and innovation in personalized cancer care, underscoring the transformative potential of multi-omics approaches in the fight against cancer.

\section*{Data Availability}
The molecular data, overall survival information, and other phenotypes from the TCGA and corresponding labels are available from NIH Genomic Data Commons (\url{https://portal.gdc.cancer.gov/}). The gene expression, miRNA expression, and DNA Methylation data was obtained from UCSC XENA (\url{https://xena.ucsc.edu/}). The CPTAC-LSCC and Moffitt LSCC data are available at \cite{UCSCXena, stewart2019proteogenomic}. The codebase for the project are available at \url{https://github.com/lab-rasool/SeNMo}.

\bibliographystyle{unsrt}  
\bibliography{references}  

\begin{thebibliography}{100}

\bibitem{jiang2022big}
Peng Jiang, Sanju Sinha, Kenneth Aldape, Sridhar Hannenhalli, Cenk Sahinalp, and Eytan Ruppin.
\newblock Big data in basic and translational cancer research.
\newblock {\em Nature Reviews Cancer}, 22(11):625--639, 2022.

\bibitem{bera2022predicting}
Kaustav Bera, Nathaniel Braman, Amit Gupta, Vamsidhar Velcheti, and Anant Madabhushi.
\newblock Predicting cancer outcomes with radiomics and artificial intelligence in radiology.
\newblock {\em Nature reviews Clinical oncology}, 19(2):132--146, 2022.

\bibitem{krithiga2021breast}
R~Krithiga and P~Geetha.
\newblock Breast cancer detection, segmentation and classification on histopathology images analysis: a systematic review.
\newblock {\em Archives of Computational Methods in Engineering}, 28(4):2607--2619, 2021.

\bibitem{morin2021artificial}
Olivier Morin, Martin Valli{\`e}res, Steve Braunstein, Jorge~Barrios Ginart, Taman Upadhaya, Henry~C Woodruff, Alex Zwanenburg, Avishek Chatterjee, Javier~E Villanueva-Meyer, Gilmer Valdes, et~al.
\newblock An artificial intelligence framework integrating longitudinal electronic health records with real-world data enables continuous pan-cancer prognostication.
\newblock {\em Nature Cancer}, 2(7):709--722, 2021.

\bibitem{chatsirisupachai2021integrative}
Kasit Chatsirisupachai, Tom Lesluyes, Luminita Paraoan, Peter Van~Loo, and Jo{\~a}o~Pedro De~Magalh{\~a}es.
\newblock An integrative analysis of the age-associated multi-omic landscape across cancers.
\newblock {\em Nature communications}, 12(1):2345, 2021.

\bibitem{hanahan2011hallmarks}
Douglas Hanahan and Robert~A Weinberg.
\newblock Hallmarks of cancer: the next generation.
\newblock {\em cell}, 144(5):646--674, 2011.

\bibitem{acosta2022multimodal}
Juli{\'a}n~N Acosta, Guido~J Falcone, Pranav Rajpurkar, and Eric~J Topol.
\newblock Multimodal biomedical ai.
\newblock {\em Nature Medicine}, 28(9):1773--1784, 2022.

\bibitem{NGS}
Dahui Qin.
\newblock Next-generation sequencing and its clinical application.
\newblock {\em Cancer biology \& medicine}, 16(1):4, 2019.

\bibitem{waqas2023multimodal}
Asim Waqas, Aakash Tripathi, Ravi~P Ramachandran, Paul Stewart, and Ghulam Rasool.
\newblock Multimodal data integration for oncology in the era of deep neural networks: a review.
\newblock {\em arXiv preprint arXiv:2303.06471}, 2023.

\bibitem{zhao2024tutorial}
Zhi Zhao, John Zobolas, Manuela Zucknick, and Tero Aittokallio.
\newblock Tutorial on survival modeling with applications to omics data.
\newblock {\em Bioinformatics}, page btae132, 2024.

\bibitem{hasin2017multi}
Yehudit Hasin, Marcus Seldin, and Aldons Lusis.
\newblock Multi-omics approaches to disease.
\newblock {\em Genome biology}, 18:1--15, 2017.

\bibitem{underwood2020pan}
Timothy Underwood.
\newblock Pan-cancer analysis of whole genomes.
\newblock {\em Nature}, 578(7793):82--93, 2020.

\bibitem{hu2020multi}
Zheng Hu, Zan Li, Zhicheng Ma, and Christina Curtis.
\newblock Multi-cancer analysis of clonality and the timing of systemic spread in paired primary tumors and metastases.
\newblock {\em Nature genetics}, 52(7):701--708, 2020.

\bibitem{sanchez2018oncogenic}
Francisco Sanchez-Vega, Marco Mina, Joshua Armenia, Walid~K Chatila, Augustin Luna, Konnor~C La, Sofia Dimitriadoy, David~L Liu, Havish~S Kantheti, Sadegh Saghafinia, et~al.
\newblock Oncogenic signaling pathways in the cancer genome atlas.
\newblock {\em Cell}, 173(2):321--337, 2018.

\bibitem{hoadley2018cell}
Katherine~A Hoadley, Christina Yau, Toshinori Hinoue, Denise~M Wolf, Alexander~J Lazar, Esther Drill, Ronglai Shen, Alison~M Taylor, Andrew~D Cherniack, V{\'e}steinn Thorsson, et~al.
\newblock Cell-of-origin patterns dominate the molecular classification of 10,000 tumors from 33 types of cancer.
\newblock {\em Cell}, 173(2):291--304, 2018.

\bibitem{thorsson2018immune}
V{\'e}steinn Thorsson, David~L Gibbs, Scott~D Brown, Denise Wolf, Dante~S Bortone, Tai-Hsien~Ou Yang, Eduard Porta-Pardo, Galen~F Gao, Christopher~L Plaisier, James~A Eddy, et~al.
\newblock The immune landscape of cancer.
\newblock {\em Immunity}, 48(4):812--830, 2018.

\bibitem{li2023pan}
Yize Li, Eduard Porta-Pardo, Collin Tokheim, Matthew~H Bailey, Tomer~M Yaron, Vasileios Stathias, Yifat Geffen, Kathleen~J Imbach, Song Cao, Shankara Anand, et~al.
\newblock Pan-cancer proteogenomics connects oncogenic drivers to functional states.
\newblock {\em Cell}, 186(18):3921--3944, 2023.

\bibitem{acharya2024comprehensive}
Debabrata Acharya and Anirban Mukhopadhyay.
\newblock A comprehensive review of machine learning techniques for multi-omics data integration: challenges and applications in precision oncology.
\newblock {\em Briefings in Functional Genomics}, page elae013, 2024.

\bibitem{ahmed2023transformers}
Sabeen Ahmed, Ian~E Nielsen, Aakash Tripathi, Shamoon Siddiqui, Ravi~P Ramachandran, and Ghulam Rasool.
\newblock Transformers in time-series analysis: A tutorial.
\newblock {\em Circuits, Systems, and Signal Processing}, 42(12):7433--7466, 2023.

\bibitem{waqas2021brain}
Asim Waqas, Dimah Dera, Ghulam Rasool, Nidhal~Carla Bouaynaya, and Hassan~M Fathallah-Shaykh.
\newblock Brain tumor segmentation and surveillance with deep artificial neural networks.
\newblock {\em Deep Learning for Biomedical Data Analysis: Techniques, Approaches, and Applications}, pages 311--350, 2021.

\bibitem{ahmed2022failure}
Sabeen Ahmed, Dimah Dera, Saud~Ul Hassan, Nidhal Bouaynaya, and Ghulam Rasool.
\newblock Failure detection in deep neural networks for medical imaging.
\newblock {\em Frontiers in Medical Technology}, 4:919046, 2022.

\bibitem{waqas2022exploring}
Asim Waqas, Hamza Farooq, Nidhal~C Bouaynaya, and Ghulam Rasool.
\newblock Exploring robust architectures for deep artificial neural networks.
\newblock {\em Communications Engineering}, 1(1):46, 2022.

\bibitem{lipkova2022artificial}
Jana Lipkova, Richard~J Chen, Bowen Chen, Ming~Y Lu, Matteo Barbieri, Daniel Shao, Anurag~J Vaidya, Chengkuan Chen, Luoting Zhuang, Drew~FK Williamson, et~al.
\newblock Artificial intelligence for multimodal data integration in oncology.
\newblock {\em Cancer cell}, 40(10):1095--1110, 2022.

\bibitem{boehm2022harnessing}
Kevin~M Boehm, Pegah Khosravi, Rami Vanguri, Jianjiong Gao, and Sohrab~P Shah.
\newblock Harnessing multimodal data integration to advance precision oncology.
\newblock {\em Nature Reviews Cancer}, 22(2):114--126, 2022.

\bibitem{he2023artificial}
Xiujing He, Xiaowei Liu, Fengli Zuo, Hubing Shi, and Jing Jing.
\newblock Artificial intelligence-based multi-omics analysis fuels cancer precision medicine.
\newblock In {\em Seminars in Cancer Biology}, volume~88, pages 187--200. Elsevier, 2023.

\bibitem{steyaert2023multimodal}
Sandra Steyaert, Marija Pizurica, Divya Nagaraj, Priya Khandelwal, Tina Hernandez-Boussard, Andrew~J Gentles, and Olivier Gevaert.
\newblock Multimodal data fusion for cancer biomarker discovery with deep learning.
\newblock {\em Nature machine intelligence}, 5(4):351--362, 2023.

\bibitem{waqas2024bio24}
Asim Waqas, Aakash Tripathi, Ashwin Mukund, Paul Stewart, Mia Naeini, and Ghulam Rasool.
\newblock Bio24-031: Hierarchical multimodal learning on pan-squamous cell carcinomas for improved survival outcomes.
\newblock {\em Journal of the National Comprehensive Cancer Network}, 22(2.5), 2024.

\bibitem{tripathi2024multimodal}
Aakash Tripathi, Asim Waqas, Yasin Yilmaz, and Ghulam Rasool.
\newblock Multimodal transformer model improves survival prediction in lung cancer compared to unimodal approaches.
\newblock {\em Cancer Research}, 84(6\_Supplement):4905--4905, 2024.

\bibitem{MINDS}
Aakash Tripathi, Asim Waqas, Kavya Venkatesan, Yasin Yilmaz, and Ghulam Rasool.
\newblock Building flexible, scalable, and machine learning-ready multimodal oncology datasets.
\newblock {\em Sensors}, 24(5):1634, 2024.

\bibitem{li2020pan}
Junyi Li, Qingzhe Xu, Mingxiao Wu, Tao Huang, and Yadong Wang.
\newblock Pan-cancer classification based on self-normalizing neural networks and feature selection.
\newblock {\em Frontiers in Bioengineering and Biotechnology}, 8:766, 2020.

\bibitem{chen2022pan}
Richard~J Chen, Ming~Y Lu, Drew~FK Williamson, Tiffany~Y Chen, Jana Lipkova, Zahra Noor, Muhammad Shaban, Maha Shady, Mane Williams, Bumjin Joo, et~al.
\newblock Pan-cancer integrative histology-genomic analysis via multimodal deep learning.
\newblock {\em Cancer Cell}, 40(8):865--878, 2022.

\bibitem{poirion2021deepprog}
Olivier~B Poirion, Zheng Jing, Kumardeep Chaudhary, Sijia Huang, and Lana~X Garmire.
\newblock Deepprog: an ensemble of deep-learning and machine-learning models for prognosis prediction using multi-omics data.
\newblock {\em Genome medicine}, 13:1--15, 2021.

\bibitem{khadirnaikar2023integration}
Seema Khadirnaikar, Sudhanshu Shukla, and SRM Prasanna.
\newblock Integration of pan-cancer multi-omics data for novel mixed subgroup identification using machine learning methods.
\newblock {\em Plos one}, 18(10):e0287176, 2023.

\bibitem{ma2019integrate}
Tianle Ma and Aidong Zhang.
\newblock Integrate multi-omics data with biological interaction networks using multi-view factorization autoencoder (mae).
\newblock {\em BMC genomics}, 20(Suppl 11):944, 2019.

\bibitem{zhao2020identification}
Ning Zhao, Maozu Guo, Kuanquan Wang, Chunlong Zhang, and Xiaoyan Liu.
\newblock Identification of pan-cancer prognostic biomarkers through integration of multi-omics data.
\newblock {\em Frontiers in Bioengineering and Biotechnology}, 8:268, 2020.

\bibitem{ellen2023autoencoder}
Jacob~G Ellen, Etai Jacob, Nikos Nikolaou, and Natasha Markuzon.
\newblock Autoencoder-based multimodal prediction of non-small cell lung cancer survival.
\newblock {\em Scientific Reports}, 13(1):15761, 2023.

\bibitem{nikolaou2024quantifying}
Nikolaos Nikolaou, Domingo Salazar, Harish RaviPrakash, Miguel Goncalves, Rob Mulla, Nikolay Burlutskiy, Natasha Markuzon, and Etai Jacob.
\newblock Quantifying the advantage of multimodal data fusion for survival prediction in cancer patients.
\newblock {\em bioRxiv}, pages 2024--01, 2024.

\bibitem{rong2022mcluster}
Zhiwei Rong, Zhilin Liu, Jiali Song, Lei Cao, Yipe Yu, Mantang Qiu, and Yan Hou.
\newblock Mcluster-vaes: an end-to-end variational deep learning-based clustering method for subtype discovery using multi-omics data.
\newblock {\em Computers in Biology and Medicine}, 150:106085, 2022.

\bibitem{pan2023multi}
Liangrui Pan, Dazhen Liu, Yutao Dou, Lian Wang, Zhichao Feng, Pengfei Rong, Liwen Xu, and Shaoliang Peng.
\newblock Multi-head attention mechanism learning for cancer new subtypes and treatment based on cancer multi-omics data.
\newblock {\em arXiv preprint arXiv:2307.04075}, 2023.

\bibitem{jia2022feature}
Weikuan Jia, Meili Sun, Jian Lian, and Sujuan Hou.
\newblock Feature dimensionality reduction: a review.
\newblock {\em Complex \& Intelligent Systems}, 8(3):2663--2693, 2022.

\bibitem{krawczuk2016feature}
Jerzy Krawczuk and Tomasz {\L}ukaszuk.
\newblock The feature selection bias problem in relation to high-dimensional gene data.
\newblock {\em Artificial intelligence in medicine}, 66:63--71, 2016.

\bibitem{yang2023causal}
Shuai Yang, Xianjie Guo, Kui Yu, Xiaoling Huang, Tingting Jiang, Jin He, and Lichuan Gu.
\newblock Causal feature selection in the presence of sample selection bias.
\newblock {\em ACM Transactions on Intelligent Systems and Technology}, 14(5):1--18, 2023.

\bibitem{waqas2023revolutionizing}
Asim Waqas, Marilyn~M Bui, Eric~F Glassy, Issam El~Naqa, Piotr Borkowski, Andrew~A Borkowski, and Ghulam Rasool.
\newblock Revolutionizing digital pathology with the power of generative artificial intelligence and foundation models.
\newblock {\em Laboratory Investigation}, page 100255, 2023.

\bibitem{hartsock2024vision}
Iryna Hartsock and Ghulam Rasool.
\newblock Vision-language models for medical report generation and visual question answering: A review.
\newblock {\em arXiv preprint arXiv:2403.02469}, 2024.

\bibitem{brown2020language}
Tom Brown, Benjamin Mann, Nick Ryder, Melanie Subbiah, Jared~D Kaplan, Prafulla Dhariwal, Arvind Neelakantan, Pranav Shyam, Girish Sastry, Amanda Askell, et~al.
\newblock Language models are few-shot learners.
\newblock {\em Advances in neural information processing systems}, 33:1877--1901, 2020.

\bibitem{devlin2018bert}
Jacob Devlin, Ming-Wei Chang, Kenton Lee, and Kristina Toutanova.
\newblock Bert: Pre-training of deep bidirectional transformers for language understanding.
\newblock {\em arXiv preprint arXiv:1810.04805}, 2018.

\bibitem{radford2021learning}
Alec Radford, Jong~Wook Kim, Chris Hallacy, Aditya Ramesh, Gabriel Goh, Sandhini Agarwal, Girish Sastry, Amanda Askell, Pamela Mishkin, Jack Clark, et~al.
\newblock Learning transferable visual models from natural language supervision.
\newblock In {\em International conference on machine learning}, pages 8748--8763. PMLR, 2021.

\bibitem{lu2019vilbert}
Jiasen Lu, Dhruv Batra, Devi Parikh, and Stefan Lee.
\newblock Vilbert: Pretraining task-agnostic visiolinguistic representations for vision-and-language tasks.
\newblock {\em Advances in neural information processing systems}, 32, 2019.

\bibitem{TCGA}
Katarzyna Tomczak, Patrycja Czerwi{\'n}ska, and Maciej Wiznerowicz.
\newblock {Review The Cancer Genome Atlas (TCGA): An immeasurable source of knowledge}.
\newblock {\em Contemporary Oncology}, 2015(1):68--77, 2015.

\bibitem{CPTAC}
Matthew~J. Ellis, Michael Gillette, Steven~A. Carr, Amanda~G. Paulovich, Richard~D. Smith, Karin~K. Rodland, R.~Reid Townsend, Christopher Kinsinger, Mehdi Mesri, Henry Rodriguez, and Daniel~C. Liebler.
\newblock {Connecting Genomic Alterations to Cancer Biology with Proteomics: The NCI Clinical Proteomic Tumor Analysis Consortium}.
\newblock {\em Cancer Discovery}, 3(10):1108--1112, 10 2013.

\bibitem{cui2024scgpt}
Haotian Cui, Chloe Wang, Hassaan Maan, Kuan Pang, Fengning Luo, Nan Duan, and Bo~Wang.
\newblock scgpt: toward building a foundation model for single-cell multi-omics using generative ai.
\newblock {\em Nature Methods}, pages 1--11, 2024.

\bibitem{zhu2023samms}
Wen Zhu, Yiwen Chen, Shanling Nie, and Hai Yang.
\newblock Samms: Multi-modality deep learning with the foundation model for the prediction of cancer patient survival.
\newblock In {\em 2023 IEEE International Conference on Bioinformatics and Biomedicine (BIBM)}, pages 3662--3668. IEEE, 2023.

\bibitem{chen2022interpretable}
Jiayang Chen, Zhihang Hu, Siqi Sun, Qingxiong Tan, Yixuan Wang, Qinze Yu, Licheng Zong, Liang Hong, Jin Xiao, Tao Shen, et~al.
\newblock Interpretable rna foundation model from unannotated data for highly accurate rna structure and function predictions.
\newblock {\em arXiv preprint arXiv:2204.00300}, 2022.

\bibitem{wang2024path}
Hongxiao Wang, Yang Yang, Zhuo Zhao, Pengfei Gu, Nishchal Sapkota, and Danny~Z Chen.
\newblock Path-gptomic: A balanced multi-modal learning framework for survival outcome prediction.
\newblock {\em arXiv preprint arXiv:2403.11375}, 2024.

\bibitem{alfasly2023foundation}
Saghir Alfasly, Peyman Nejat, Sobhan Hemati, Jibran Khan, Isaiah Lahr, Areej Alsaafin, Abubakr Shafique, Nneka Comfere, Dennis Murphree, Chady Meroueh, et~al.
\newblock When is a foundation model a foundation model.
\newblock {\em arXiv preprint arXiv:2309.11510}, 2023.

\bibitem{UCSCXena}
Mary Goldman, Brian Craft, Mim Hastie, Kristupas Repe{\v{c}}ka, Fran McDade, Akhil Kamath, Ayan Banerjee, Yunhai Luo, Dave Rogers, Angela~N Brooks, et~al.
\newblock The ucsc xena platform for public and private cancer genomics data visualization and interpretation.
\newblock {\em biorxiv}, page 326470, 2018.

\bibitem{CPTAC-LSCC}
Shankha Satpathy, Karsten Krug, Pierre M~Jean Beltran, Sara~R Savage, Francesca Petralia, Chandan Kumar-Sinha, Yongchao Dou, Boris Reva, M~Harry Kane, Shayan~C Avanessian, et~al.
\newblock A proteogenomic portrait of lung squamous cell carcinoma.
\newblock {\em Cell}, 184(16):4348--4371, 2021.

\bibitem{stewart2019proteogenomic}
Paul~A Stewart, Eric~A Welsh, Robbert~JC Slebos, Bin Fang, Victoria Izumi, Matthew Chambers, Guolin Zhang, Ling Cen, Fredrik Pettersson, Yonghong Zhang, et~al.
\newblock Proteogenomic landscape of squamous cell lung cancer.
\newblock {\em Nature communications}, 10(1):3578, 2019.

\bibitem{sarhadi2022molecular}
Virinder~Kaur Sarhadi and Gemma Armengol.
\newblock Molecular biomarkers in cancer.
\newblock {\em Biomolecules}, 12(8):1021, 2022.

\bibitem{chen2021moving}
Feng Chen, Michael~C Wendl, Matthew~A Wyczalkowski, Matthew~H Bailey, Yize Li, and Li~Ding.
\newblock Moving pan-cancer studies from basic research toward the clinic.
\newblock {\em Nature cancer}, 2(9):879--890, 2021.

\bibitem{loyfer2023dna}
Netanel Loyfer, Judith Magenheim, Ayelet Peretz, Gordon Cann, Joerg Bredno, Agnes Klochendler, Ilana Fox-Fisher, Sapir Shabi-Porat, Merav Hecht, Tsuria Pelet, et~al.
\newblock A dna methylation atlas of normal human cell types.
\newblock {\em Nature}, 613(7943):355--364, 2023.

\bibitem{lakshminarasimhan2016role}
Ranjani Lakshminarasimhan and Gangning Liang.
\newblock The role of dna methylation in cancer.
\newblock {\em DNA Methyltransferases-Role and Function}, pages 151--172, 2016.

\bibitem{du2010comparison}
Pan Du, Xiao Zhang, Chiang-Ching Huang, Nadereh Jafari, Warren~A Kibbe, Lifang Hou, and Simon~M Lin.
\newblock Comparison of beta-value and m-value methods for quantifying methylation levels by microarray analysis.
\newblock {\em BMC bioinformatics}, 11:1--9, 2010.

\bibitem{wang2018framework}
Zhenxing Wang, XiaoLiang Wu, and Yadong Wang.
\newblock A framework for analyzing dna methylation data from illumina infinium humanmethylation450 beadchip.
\newblock {\em BMC bioinformatics}, 19:15--22, 2018.

\bibitem{corchete2020systematic}
Luis~A Corchete, Elizabeta~A Rojas, Diego Alonso-L{\'o}pez, Javier De~Las~Rivas, Norma~C Guti{\'e}rrez, and Francisco~J Burguillo.
\newblock Systematic comparison and assessment of rna-seq procedures for gene expression quantitative analysis.
\newblock {\em Scientific reports}, 10(1):19737, 2020.

\bibitem{hijazo2021gene}
Sara Hijazo-Pechero, Ania Alay, Ra{\'u}l Mar{\'\i}n, Noelia Vilari{\~n}o, Cristina Mu{\~n}oz-Pinedo, Alberto Villanueva, David Santamar{\'\i}a, Ernest Nadal, and Xavier Sol{\'e}.
\newblock Gene expression profiling as a potential tool for precision oncology in non-small cell lung cancer.
\newblock {\em Cancers}, 13(19):4734, 2021.

\bibitem{gonzalez2023gene}
Augusto Gonzalez, Dario~A Leon, Yasser Perera, and Rolando Perez.
\newblock On the gene expression landscape of cancer.
\newblock {\em Plos one}, 18(2):e0277786, 2023.

\bibitem{rau2019exploring}
Andrea Rau, Michael Flister, Hallgeir Rui, and Paul~L Auer.
\newblock Exploring drivers of gene expression in the cancer genome atlas.
\newblock {\em Bioinformatics}, 35(1):62--68, 2019.

\bibitem{genexpression}
EBI Gene~Expression Team.
\newblock Expression atlas.
\newblock Software available from https://www.ebi.ac.uk.

\bibitem{peng2016role}
Yong Peng and Carlo~M Croce.
\newblock The role of micrornas in human cancer.
\newblock {\em Signal transduction and targeted therapy}, 1(1):1--9, 2016.

\bibitem{chu2016large}
Andy Chu, Gordon Robertson, Denise Brooks, Andrew~J Mungall, Inanc Birol, Robin Coope, Yussanne Ma, Steven Jones, and Marco~A Marra.
\newblock Large-scale profiling of micrornas for the cancer genome atlas.
\newblock {\em Nucleic acids research}, 44(1):e3--e3, 2016.

\bibitem{lin2022integrative}
Shuting Lin, Jie Zhou, Yiqiong Xiao, Bridget Neary, Yong Teng, and Peng Qiu.
\newblock Integrative analysis of tcga data identifies mirnas as drug-specific survival biomarkers.
\newblock {\em Scientific Reports}, 12(1):6785, 2022.

\bibitem{GDCDocs}
GDC Documentation.
\newblock Reverse phase protein array.
\newblock \url{https://docs.gdc.cancer.gov/Data/Bioinformatics_Pipelines/RPPA_intro/}, 2024.
\newblock Accessed: 2024-05-13.

\bibitem{RPPAdescription}
MD~Anderson.
\newblock Rppa description.
\newblock \url{https://www.mdanderson.org/documents/core-facilities/Functional Proteomics RPPA Core Facility/RPPA Description_2016.pdf}, 2024.
\newblock Accessed: 2024-05-13.

\bibitem{chen2019tcpa}
Mei-Ju~May Chen, Jun Li, Yumeng Wang, Rehan Akbani, Yiling Lu, Gordon~B Mills, and Han Liang.
\newblock Tcpa v3. 0: an integrative platform to explore the pan-cancer analysis of functional proteomic data.
\newblock {\em Molecular \& Cellular Proteomics}, 18(8):S15--S25, 2019.

\bibitem{li2013tcpa}
Jun Li, Yiling Lu, Rehan Akbani, Zhenlin Ju, Paul~L Roebuck, Wenbin Liu, Ji-Yeon Yang, Bradley~M Broom, Roeland~GW Verhaak, David~W Kane, et~al.
\newblock Tcpa: a resource for cancer functional proteomics data.
\newblock {\em Nature methods}, 10(11):1046--1047, 2013.

\bibitem{SuperCurve}
Zhenlin Ju, Wenbin Liu, Paul~L Roebuck, Doris~R Siwak, Nianxiang Zhang, Yiling Lu, Michael~A Davies, Rehan Akbani, John~N Weinstein, Gordon~B Mills, et~al.
\newblock Development of a robust classifier for quality control of reverse-phase protein arrays.
\newblock {\em Bioinformatics}, 31(6):912--918, 2015.

\bibitem{dnamut3}
Genomic~Data Commons.
\newblock Mutation annotation format.
\newblock \url{https://docs.gdc.cancer.gov/Encyclopedia/pages/Mutation_Annotation_Format/}, 2024.
\newblock Accessed: 2024-05-13.

\bibitem{dnamut1}
Genomic~Data Commons.
\newblock File format - vcf.
\newblock \url{https://docs.gdc.cancer.gov/Data/File_Formats/VCF_Format/}, 2024.
\newblock Accessed: 2024-05-13.

\bibitem{dnamut2}
Genomic~Data Commons.
\newblock File format - maf.
\newblock \url{https://docs.gdc.cancer.gov/Data/File_Formats/MAF_Format/}, 2024.
\newblock Accessed: 2024-05-13.

\bibitem{mendiratta2021cancer}
Gaurav Mendiratta, Eugene Ke, Meraj Aziz, David Liarakos, Melinda Tong, and Edward~C Stites.
\newblock Cancer gene mutation frequencies for the us population.
\newblock {\em Nature communications}, 12(1):5961, 2021.

\bibitem{lewandowska2022risk}
Anna Lewandowska, Grzegorz Rudzki, Tomasz Lewandowski, Aleksandra Stryjkowska-Gora, and S{\l}awomir Rudzki.
\newblock Risk factors for the diagnosis of colorectal cancer.
\newblock {\em Cancer Control}, 29:10732748211056692, 2022.

\bibitem{lopes2020genome}
Camila~M Lopes-Ramos, John Quackenbush, and Dawn~L DeMeo.
\newblock Genome-wide sex and gender differences in cancer.
\newblock {\em Frontiers in oncology}, 10:597788, 2020.

\bibitem{zavala2021cancer}
Valentina~A Zavala, Paige~M Bracci, John~M Carethers, Luis Carvajal-Carmona, Nicole~B Coggins, Marcia~R Cruz-Correa, Melissa Davis, Adam~J de~Smith, Julie Dutil, Jane~C Figueiredo, et~al.
\newblock Cancer health disparities in racial/ethnic minorities in the united states.
\newblock {\em British journal of cancer}, 124(2):315--332, 2021.

\bibitem{yang2022research}
Xinyu Yang, Dongmei Mu, Hao Peng, Hua Li, Ying Wang, Ping Wang, Yue Wang, and Siqi Han.
\newblock Research and application of artificial intelligence based on electronic health records of patients with cancer: systematic review.
\newblock {\em JMIR Medical Informatics}, 10(4):e33799, 2022.

\bibitem{liao2007logistic}
JG~Liao and Khew-Voon Chin.
\newblock Logistic regression for disease classification using microarray data: model selection in a large p and small n case.
\newblock {\em Bioinformatics}, 23(15):1945--1951, 2007.

\bibitem{zhao2021tpm}
Yingdong Zhao, Ming-Chung Li, Mariam~M Konat{\'e}, Li~Chen, Biswajit Das, Chris Karlovich, P~Mickey Williams, Yvonne~A Evrard, James~H Doroshow, and Lisa~M McShane.
\newblock Tpm, fpkm, or normalized counts? a comparative study of quantification measures for the analysis of rna-seq data from the nci patient-derived models repository.
\newblock {\em Journal of translational medicine}, 19(1):269, 2021.

\bibitem{kaushik2014spatial}
Poorvi Kaushik, Evan~J Molinelli, Martin~L Miller, Weiqing Wang, Anil Korkut, Wenbin Liu, Zhenlin Ju, Yiling Lu, Gordon Mills, and Chris Sander.
\newblock Spatial normalization of reverse phase protein array data.
\newblock {\em PloS one}, 9(12):e97213, 2014.

\bibitem{liu2014comprehensive}
Wenbin Liu, Zhenlin Ju, Yiling Lu, Gordon~B Mills, and Rehan Akbani.
\newblock A comprehensive comparison of normalization methods for loading control and variance stabilization of reverse-phase protein array data.
\newblock {\em Cancer informatics}, 13:CIN--S13329, 2014.

\bibitem{song2020review}
Meng Song, Jonathan Greenbaum, Joseph Luttrell~IV, Weihua Zhou, Chong Wu, Hui Shen, Ping Gong, Chaoyang Zhang, and Hong-Wen Deng.
\newblock A review of integrative imputation for multi-omics datasets.
\newblock {\em Frontiers in Genetics}, 11:570255, 2020.

\bibitem{anowar2021conceptual}
F~Anowar, S~Sadaoui, and B~Selim.
\newblock Conceptual and empirical comparison of dimensionality reduction algorithms (pca, kpca, lda, mds, svd, lle, isomap, le, ica, t-sne), comput. sci. rev., 40, 100378.
\newblock {\em ISI}, 2021.

\bibitem{settino2018survey}
Marzia Settino and Mario Cannataro.
\newblock Survey of main tools for querying and analyzing tcga data.
\newblock In {\em 2018 IEEE International Conference on Bioinformatics and Biomedicine (BIBM)}, pages 1711--1718. IEEE, 2018.

\bibitem{lei2023tcga}
Brian Lei, Xinyin Jiang, and Anjana Saxena.
\newblock Tcga expression analyses of 10 carcinoma types reveal clinically significant racial differences.
\newblock {\em Cancers}, 15(10):2695, 2023.

\bibitem{feture-engine}
Feature-engine, a python library for feature engineering and selection.

\bibitem{bommert2022benchmark}
Andrea Bommert, Thomas Welchowski, Matthias Schmid, and J{\"o}rg Rahnenf{\"u}hrer.
\newblock Benchmark of filter methods for feature selection in high-dimensional gene expression survival data.
\newblock {\em Briefings in Bioinformatics}, 23(1):bbab354, 2022.

\bibitem{scikit-learn}
F.~Pedregosa, G.~Varoquaux, A.~Gramfort, V.~Michel, B.~Thirion, O.~Grisel, M.~Blondel, P.~Prettenhofer, R.~Weiss, V.~Dubourg, J.~Vanderplas, A.~Passos, D.~Cournapeau, M.~Brucher, M.~Perrot, and E.~Duchesnay.
\newblock Scikit-learn: Machine learning in {P}ython.
\newblock {\em Journal of Machine Learning Research}, 12:2825--2830, 2011.

\bibitem{anggraeny2018analysis}
Fetty~Tri Anggraeny, Intan~Yuniar Purbasari, M~Syahrul Munir, Faisal Muttaqin, Eka~Prakarsa Mandyarta, and Fawwaz~Ali Akbar.
\newblock Analysis of simple data imputation in disease dataset.
\newblock In {\em International Conference on Science and Technology (ICST 2018)}, pages 471--475. Atlantis Press, 2018.

\bibitem{ulriksborg2022imputation}
Tomas~Rakv{\aa}g Ulriksborg.
\newblock Imputation of missing time series values using statistical and mathematical strategies.
\newblock {\em Department of Informatics}, 2022.

\bibitem{talwar2018autoimpute}
Divyanshu Talwar, Aanchal Mongia, Debarka Sengupta, and Angshul Majumdar.
\newblock Autoimpute: Autoencoder based imputation of single-cell rna-seq data.
\newblock {\em Scientific reports}, 8(1):16329, 2018.

\bibitem{yi2019not}
Joonyoung Yi, Juhyuk Lee, Kwang~Joon Kim, Sung~Ju Hwang, and Eunho Yang.
\newblock Why not to use zero imputation? correcting sparsity bias in training neural networks.
\newblock {\em arXiv preprint arXiv:1906.00150}, 2019.

\bibitem{chen2020pathomic}
Richard~J Chen, Ming~Y Lu, Jingwen Wang, Drew~FK Williamson, Scott~J Rodig, Neal~I Lindeman, and Faisal Mahmood.
\newblock Pathomic fusion: an integrated framework for fusing histopathology and genomic features for cancer diagnosis and prognosis.
\newblock {\em IEEE Transactions on Medical Imaging}, 41(4):757--770, 2020.

\bibitem{patwardhan2024towards}
Kedar~A Patwardhan, Harish RaviPrakash, Nikos Nikolaou, Ignacio Gonzalez-Garcia, Jose~Domingo Salazar, Paul Metcalfe, and Joachim Reischl.
\newblock Towards a survival risk prediction model for metastatic nsclc patients on durvalumab using whole-lung ct radiomics.
\newblock {\em bioRxiv}, pages 2024--02, 2024.

\bibitem{miller2019cancer}
Kimberly~D Miller, Leticia Nogueira, Angela~B Mariotto, Julia~H Rowland, K~Robin Yabroff, Catherine~M Alfano, Ahmedin Jemal, Joan~L Kramer, and Rebecca~L Siegel.
\newblock Cancer treatment and survivorship statistics, 2019.
\newblock {\em CA: a cancer journal for clinicians}, 69(5):363--385, 2019.

\bibitem{van2024TLS}
Mart van Rijthoven, Simon Obahor, Fabio Pagliarulo, Maries van~den Broek, Peter Schraml, Holger Moch, Jeroen van~der Laak, Francesco Ciompi, and Karina Silina.
\newblock Multi-resolution deep learning characterizes tertiary lymphoid structures and their prognostic relevance in solid tumors.
\newblock {\em Communications Medicine}, 4(1):5, 2024.

\bibitem{chen2024TLS}
Ziqiang Chen, Xiaobing Wang, Zelin Jin, Bosen Li, Dongxian Jiang, Yanqiu Wang, Mengping Jiang, Dandan Zhang, Pei Yuan, Yahui Zhao, et~al.
\newblock Deep learning on tertiary lymphoid structures in hematoxylin-eosin predicts cancer prognosis and immunotherapy response.
\newblock {\em NPJ Precision Oncology}, 8(1):73, 2024.

\bibitem{klambauer2017self}
G{\"u}nter Klambauer, Thomas Unterthiner, Andreas Mayr, and Sepp Hochreiter.
\newblock Self-normalizing neural networks.
\newblock {\em Advances in neural information processing systems}, 30, 2017.

\bibitem{wandb}
Lukas Biewald.
\newblock Experiment tracking with weights and biases, 2020.
\newblock Software available from wandb.com.

\bibitem{Cox-nnet}
Travers Ching.
\newblock Cox regression.
\newblock \url{http://traversc.github.io/cox-nnet/docs/}, 2024.
\newblock Accessed: 2024-05-13.

\bibitem{Lifelines}
Cameron Davidson-Pilon.
\newblock lifelines, survival analysis in python.
\newblock \url{https://doi.org/10.5281/zenodo.10456828}, Jan 2024.
\newblock Accessed: 2024-05-13.

\bibitem{Huber}
PyTorch Documentation.
\newblock Huberloss.
\newblock \url{https://pytorch.org/docs/stable/generated/torch.nn.HuberLoss.html}, 2024.
\newblock Accessed: 2024-10-24.

\bibitem{li2023survival}
Zhe Li, Yuming Jiang, Mengkang Lu, Ruijiang Li, and Yong Xia.
\newblock Survival prediction via hierarchical multimodal co-attention transformer: A computational histology-radiology solution.
\newblock {\em IEEE Transactions on Medical Imaging}, 2023.

\bibitem{liu2024kan}
Ziming Liu, Yixuan Wang, Sachin Vaidya, Fabian Ruehle, James Halverson, Marin Solja{\v{c}}i{\'c}, Thomas~Y Hou, and Max Tegmark.
\newblock Kan: Kolmogorov-arnold networks.
\newblock {\em arXiv preprint arXiv:2404.19756}, 2024.

\bibitem{gore2022cancernet}
Steven Gore and Rajeev~K Azad.
\newblock Cancernet: a unified deep learning network for pan-cancer diagnostics.
\newblock {\em BMC bioinformatics}, 23(1):229, 2022.

\bibitem{tripathi2024honeybee}
Aakash Tripathi, Asim Waqas, Yasin Yilmaz, and Ghulam Rasool.
\newblock Honeybee: A scalable modular framework for creating multimodal oncology datasets with foundational embedding models.
\newblock {\em arXiv preprint arXiv:2405.07460}, 2024.

\end{thebibliography}

\section*{Appendix A1: Hyperparameters Search - Training on Pan-cancer Multiomics Data}\label{secA1}



\end{document}